\documentclass{article} 
\usepackage{iclr2026_conference,times}

\usepackage{amsmath,amsfonts,bm}

\def\eqref#1{equation~\ref{#1}}

\def\1{\bm{1}}

\DeclareMathAlphabet{\mathsfit}{\encodingdefault}{\sfdefault}{m}{sl}
\SetMathAlphabet{\mathsfit}{bold}{\encodingdefault}{\sfdefault}{bx}{n}

\newcommand{\Bl}{\hat{B}^{l}}
\newcommand{\Bu}{\hat{B}^{u}}
\newcommand{\rl}{\hat{r}^{l}}
\newcommand{\ru}{\hat{r}^{u}}
\newcommand{\Ll}{L^{l}}
\newcommand{\Lu}{L^{u}}
\newcommand{\Ul}{U^{l}}
\newcommand{\Uu}{U^{u}}
\newcommand{\Cll}{\hat{C}^{\text{ll}}}
\newcommand{\Clu}{\hat{C}^{\text{lu}}}
\newcommand{\Cul}{\hat{C}^{\text{ul}}}
\newcommand{\Cuu}{\hat{C}^{\text{uu}}}
\newcommand{\tll}{\tau^{\text{ll}}}
\newcommand{\tlu}{\tau^{\text{lu}}}
\newcommand{\tul}{\tau^{\text{ul}}}
\newcommand{\tuu}{\tau^{\text{uu}}}

\usepackage[hyphens]{url}  
\usepackage{graphicx} 
\usepackage{subcaption}
\usepackage{hyperref}       
\usepackage{url}            
\usepackage{booktabs}       
\usepackage{nicefrac}       
\usepackage{microtype}      
\usepackage{xcolor}         

\usepackage[bibliography=common]{apxproof}  
\usepackage{array} 
\usepackage{float} 
\usepackage[normalem]{ulem}
\usepackage{adjustbox}
\usepackage{multirow} 
\usepackage{threeparttable}
\usepackage{graphicx} 
\usepackage{nicefrac}
\usepackage{algorithm} 
\usepackage{algorithmic} 

\newtheoremrep{theorem}{Theorem}

\newtheoremrep{proposition}{Proposition}

\title{Tightening Optimality Gap with Confidence through Conformal Prediction}

\author{%
  Miao Li\thanks{Corresponding author.} \\
  H. Milton Stewart School of Industrial and Systems Engineering\\
  Georgia Institute of Technology\\
  Atlanta, GA 30332-0205 \\
  \texttt{mli746@gatech.edu} \\
  \And
  Michael Klamkin \\
  H. Milton Stewart School of Industrial and Systems Engineering\\
  Georgia Institute of Technology\\
  Atlanta, GA 30332-0205 \\
  \texttt{mklamkin3@gatech.edu} \\ 
   \And
   Russell Bent\\
   Los Alamos National Laboratory\\
   Los Alamos, NM 87545\\
   \texttt{rbent@lanl.gov}\\
  \And
  Pascal {Van Hentenryck} \\
  H. Milton Stewart School of Industrial and Systems Engineering\\
  Georgia Institute of Technology\\
  Atlanta, GA 30332-0205 \\
\texttt{phentenryck3@gatech.edu} \\ 
}

\iclrfinalcopy 
\begin{document}

\maketitle
\begin{abstract}
Decision makers routinely use constrained optimization technology to plan and operate complex systems like global supply chains or power grids. In this context, practitioners must assess how close a computed solution is to optimality in order to make operational decisions, such as whether the current solution is sufficient or whether additional computation is warranted. A common practice is to evaluate solution quality using dual bounds returned by optimization solvers. While these dual bounds come with certified guarantees, they are often too loose to be practically informative. 

To this end, this paper introduces a novel conformal prediction framework for tightening loose primal and dual bounds. The proposed method addresses the heteroskedasticity commonly observed in these bounds via selective inference, and further exploits their inherent certified validity to produce tighter, more informative prediction intervals. Finally, numerical experiments on large-scale industrial problems suggest that the proposed approach can provide the same coverage level more efficiently than baseline methods.

\end{abstract}

\section{Introduction}
\label{sec:intro}
Constrained optimization is instrumental in operating complex systems efficiently, with applications spanning supply chains, logistics and manufacturing, as well as healthcare and power systems. In these settings, operational workflows often require solving many similar problem instances under strict time constraints. For many industry-scale problems, computing near-optimal solutions for all required instances within operational deadlines is not realistic, requiring smart allocation of computational resources such that the marginal value of compute is maximized.

Assessing the adequacy of a solution often requires computing an additional \emph{dual bound}, a mathematical certificate of solution quality. Such bounds are typically obtained by solving a relaxation \cite{boyd2004convex,geoffrion2009lagrangean}, via branch-and-bound algorithms \cite{wolsey1999integer}, or through advanced machine-learning-based proxies \cite{tanneau2024dual, tordesillas2023rayen, grontas2025pinet,klamkin2025self}. In practice, solution quality is commonly monitored through the \emph{optimality gap}, i.e the distance between the incumbent (primal) objective value and the best available dual bound. However, it is widely recognized that in operational settings, dual bounds can be too loose to accurately reflect true optimality. In the worst case, an algorithm may find a near-optimal solution early on, yet the gap remains above the prescribed tolerance; the remaining computation is then spent primarily on tightening the dual bound to certify optimality by pushing it toward the incumbent objective value \cite{miltenberger2025mipgap}. This behavior can lead to inefficient resource allocation, where compute is consumed primarily to certify optimality rather than to improve the solution that practitioners ultimately care about. More broadly, in time-constrained deployments, practitioners must decide how to distribute scarce computational resources across many optimization runs. This requires estimates of the marginal gains from additional compute that are not only reliable but also tight enough to meaningfully guide early stopping and prioritization.

To address this challenge, this paper introduces a data-driven {uncertainty quantification} (UQ) method that improves the informativeness of primal--dual optimality gaps by calibrating them into tighter prediction intervals with finite-sample coverage guarantees. The key idea is to leverage conformal prediction (CP) to construct prediction intervals for the true optimal value that are both tight and equipped with statistical guarantees, thereby enabling more informed operational decisions. While certified optimality gaps via primal-dual bounds is fundamental in optimization theory, the broader use of high-probability intervals with relaxed coverage to guide decision-making is also well established in the optimization literature (e.g., in stochastic programming, where exact coverage certificates are typically unavailable \cite{mak1999monte}). Relatedly, recent work has explored learning-based approaches for estimating optimal values in constrained optimization \cite{scavuzzo2024learning,zhang2021convex,rosemberg2024learning}. These methods produce point predictions, but generally do not provide risk-controlled uncertainty quantification. In contrast, this paper seeks to refine primal--dual bounds into statistically valid, risk-controlled prediction intervals at a user-specified coverage level via CP, tailored to constrained optimization. 

To the best of the authors’ knowledge, this paper is the first to apply CP to quantify solution suboptimality in constrained optimization. 
To construct a two-sided bracket on the optimal value, the proposed CPUL-OMLT framework incorporates a novel primal-dual bound calibration procedure, providing much tighter intervals than existing approaches.

The paper’s contributions are summarized as follows:
(1) it presents a novel data-driven methodology to assess solution optimality with distribution-free, finite-sample guarantees;
(2) it introduces CPUL–OMLT, a general CP framework that is designed to maximally exploit primal and dual bounds for efficient coverage;
(3) it reports large-scale experiments on economic dispatch instances representative of real-time electricity markets.
\subsection{Problem statement}
\label{sec:intro:problem_statement}

Consider the generic constrained optimization problem of the form
{\begin{subequations}
\label{eq:optimization_generic}
\begin{align*}
    \Phi(x) = 
    \min_{s} \quad & f_{x}(s) \quad
    \text{s.t.} \quad g_{x}(s) \leq 0,
\end{align*}
\end{subequations}
}where $x \in \mathcal{X} \subseteq \mathbb{R}^{d}$ denotes the instance's parameters, $s \in \mathbb{R}^{n}$ is decision variable, $f_{x}: \mathbb{R}^{n} \rightarrow \mathbb{R}$ is the objective function to be minimized, $g_{x}: \mathbb{R}^{n} \rightarrow \mathbb{R}^{m}$ encodes constraints and $\mathcal{S}_{x} = \{g_{x}(s) \leq 0\}$ is the set of feasible solutions, and $\Phi(x)$ is the instance's optimal value.
The paper assumes the availability of a primal-dual pair in the following form.
A \emph{primal} solution $\bar{s} \in \mathcal{S}_{x}$ provides a primal (upper) bound on the optimal value, i.e., $f_{x}(\bar{s}) \geq \Phi(x)$.
Such primal solutions can be obtained using either heuristics or exact algorithms.
Conversely, a \emph{dual} (lower) bound $\psi(x): \mathbb{R}^m \rightarrow \mathbb{R}$ is guaranteed to be smaller than the optimal value, i.e., $\psi(x) \leq \Phi(x)$. This assumption is standard in practice: modern global optimization solvers maintain a certified dual bound throughout the solution process, enabling practitioners to assess solution quality (e.g., \cite{miltenberger2025mipgap}).

\paragraph{General Problem Setting.}
For simplicity and alignment with the conformal prediction literature, let $X \in \mathcal{X} \subseteq \mathbb{R}^d$ denote random instance parameters (features), and define the corresponding optimal value (label) as $Y := \phi(X)$.
Assume that functions $\Bl(X)$ and $\Bu(X)$ are available such that they form valid lower and upper bounds on $Y$, i.e.,
    $\Bl(X) \le Y \le \Bu(X)$.
These bounds induce a trivial $100\%$ prediction interval $[\Bl(X),\,\Bu(X)]$ for $Y$. For readability and alignment with the CP literature, the terms \emph{lower} and \emph{upper} bounds are used instead of \emph{dual} and \emph{primal}: in minimization, $\Bl$ and $\Bu$ correspond to dual and primal bounds respectively, and vice-versa in maximization.

\paragraph{Problem Formulation.}
The goal of this paper is to construct tighter prediction intervals for $Y$ by targeting $(1-\alpha)$ marginal coverage (with $\alpha \in [0,1]$), refining the perfect-coverage bounds $[\Bl(X),\Bu(X)]$.
Following split conformal prediction \cite{vovk2005algorithmic}, partition the $N$ i.i.d. samples into a training set $\mathcal{D}_{\text{train}}$ and a calibration set $\mathcal{D}_{\text{cal}}$, with index sets $\mathcal{I}_{\text{train}}$ and $\mathcal{I}_{\text{cal}}$.
Given a new i.i.d. test instance $X_{N+1} \sim \mathcal{P}_X$, where $\mathcal{P}_X$ denotes the marginal distribution of $X$, the goal is to refine the initial interval
   $\tilde{C}(X_{N+1}) = [\Bl(X_{N+1}),\, \Bu(X_{N+1})] $
into a tighter interval
 $\hat{C}(X_{N+1}) = [\hat{L}(X_{N+1}),\, \hat{U}(X_{N+1})]$   
that satisfies the marginal coverage guarantee
{\setlength{\abovedisplayskip}{4pt}
\setlength{\belowdisplayskip}{2pt}\begin{align}
\mathbb{P}\!\left( Y_{N+1} \in \hat{C}(X_{N+1}) \right) \ge 1-\alpha.
\end{align}}

While $\tilde{C}(X_{N+1})$ attains $100\%$ coverage by construction, it can be overly conservative; the objective is therefore to shrink interval width subject to coverage. This can be written as
{\setlength{\abovedisplayskip}{2pt}
\setlength{\belowdisplayskip}{2pt}
\begin{align}\label{eq:prediction_interval}
\min_{\hat{C}} \quad \mathbb{E}_{X \sim \mathcal{P}_X}\!\left[\, |\hat{C}(X)| \,\right] \quad
\text{s.t.} \quad \mathbb{P}\!\left( Y \in \hat{C}(X) \right) \ge 1-\alpha
\end{align}
}
\vspace{-0.3cm}
where $\hat{C}$ maps $X \in \mathbb{R}^p$ to an interval in $\mathbb{R}$.

\newcommand{\qSCP}{\hat{Q}^{\text{SCP}}}
\newcommand{\qCQR}{\hat{Q}^{\text{CQR}}}
\section{Background}
\label{sec:Background}

UQ is crucial for informed decision making, particularly in real-world applications where prediction sets are preferred over point estimates. 
Conformal Prediction, first proposed by \citet{vovk2005algorithmic}, is a widely used distribution-free UQ method, valued for its finite-sample coverage guarantees and computational efficiency. 
This section presents standard CP techniques, including Split CP, Conformal Quantile Regression, and Nested CP.
While these methods can be applied to the problem at hand (see Section~\ref{sec:experiments} and Appendix~\ref{app:experiments:cp_baselines}), unlike the proposed framework, they are not designed to explicitly leverage valid lower and upper bound predictors, hindering their efficacy in this setting.

\paragraph{Split Conformal Prediction (SCP)} 
\cite{vovk2005algorithmic,papadopoulos2002inductive,lei2018distribution} is
one of the most commonly used CP frameworks. 
Therein, a mean prediction model $\hat{f}$ is first trained using the training data $\mathcal{D}_{train}$.
SCP then produces prediction intervals of the form 
{  \setlength{\abovedisplayskip}{3pt}
  \setlength{\belowdisplayskip}{3pt}
\begin{align}
\label{eq:SCP:predict_interval}
    \hat{C}_{\text{SCP}}(x) = \left\{
        y  
        \, \middle| \,
        \qSCP_{\frac{\alpha}{2}} \leq \hat{s}(x, y) \leq \qSCP_{1-\frac{\alpha}{2}}
    \right\},
\end{align}}
where $\hat{s} \, {:} \, \mathbb{R}^{d} {\times} \mathbb{R} \, {\rightarrow} \, \mathbb{R}$ is a  \emph{conformity score}, 
and $\qSCP_{\frac{\alpha}{2}}, \qSCP_{1-\frac{\alpha}{2}}$, denote the empirical $\frac{\alpha}{2}$ and $1 {-} \frac{\alpha}{2}$ quantiles of the scores \(\{\hat{s}(X_i, Y_i)\}_{i \in  \mathcal{I}_{\text{cal}}}\).
Under the exchangeability assumption, the resulting prediction intervals have valid marginal coverage \cite{vovk2005algorithmic}, i.e.,
{\setlength{\abovedisplayskip}{2pt}
  \setlength{\belowdisplayskip}{2pt}
\begin{align*}
\label{eq:SCP:valid_coverage}
    \mathbb{P}_{X_{N+1} \sim \mathcal{P}} \left(Y_{N+1} \in \tilde{C}_{\text{SCP}}(X_{N+1})\right) \geq 1 - \alpha.
\end{align*}
}
Common choices of conformity scores include the residual score $\hat{s}(x, y) \, {=} \, y {-} \hat{f}(x)$ and the absolute residual score $\hat{s}(x, y) \, {=} \, |y {-} \hat{f}(x)|$; the reader is referred to \cite{angelopoulos2021gentle,oliveira2024split} for a more exhaustive review of SCP and conformity scores.
The SCP methodology can be applied to the paper's setting by replacing $\hat{f}$ in the above construction with either the lower ($\Bl$) or upper bound ($\Bu$) predictor.
However, this fails to jointly utilize information from both bounds.

\paragraph{Conformal Quantile Regression (CQR)}
To alleviate SCP's lack of local adaptivity, CQR \cite{romano2019conformalized}
combines quantile regression models with a conformalization procedure: First, CQR trains quantile regression models to predict the quantiles $\alpha_{\text{lo}}$ and $\alpha_{\text{hi}}$.
Then, conformalize with
$\hat{C}^{\text{CQR}}(x) = 
\left[
\hat{q}_{\alpha_{\text{lo}}}(x) {-} \qCQR_{1-\alpha}
    , \ 
\hat{q}_{\alpha_{\text{hi}}}(x) {+} \qCQR_{1-\alpha}
\right],$ 
where $\hat{q}_{\alpha_{\text{lo}}}(x), \hat{q}_{\alpha_{\text{hi}}}(x)$ are the predicted lower and upper quantiles given input data $x$, and $\qCQR_{1-\alpha}$ is the $1{-}\alpha$ quantile of the conformity scores
$\{\hat{s}^{\text{CQR}}(X_i, Y_i)\}_{i \in  \mathcal{I}_{\text{cal}}}$, 
defined as
    $\hat{s}^{\text{CQR}}(x, y) = \max \left(
        \hat{q}_{\alpha_{\text{lo}}}(x) {-} y
        , \ 
        y {-} \hat{q}_{\alpha_{\text{hi}}}(x)
    \right)$.
The CQR method could be adapted to the setting here by approximating $\hat{q}_{\alpha_{\text{lo}}}, \hat{q}_{\alpha_{\text{hi}}}$ with $\Bl, \Bu$ in the above derivation. 
In contrast to standard CQR, which relies on auxiliary quantile regression models and whose performance is sensitive to their estimation accuracy (often requiring large training datasets in high-dimensional regimes), the bounds $\Bl$ and $\Bu$ obtained from the constrained optimization formulation offer several key advantages. These bounds are often directly computable from the base model (e.g., once a feasible primal model is constructed, an upper bound can be computed by evaluating the objective at that solution), making them available at minimal cost. 
They also inherently guarantee perfect coverage, regardless of training data size. This paper focuses on effectively leveraging these readily available and valid bounds.

\paragraph{Nested Conformal Prediction (NCP)}
\cite{gupta2022nested} introduces
a unifying CP framework by utilizing nested prediction sets. 
{NCP  considers a family of nested prediction intervals }
\(\{\hat{C}_{t}\}_{t \in \mathcal{T} \subseteq \mathbb{R}}\), i.e., 
    $\forall x \in \mathcal{X}, \forall t \leq t', \hat{C}_{t}(x) \subseteq \hat{C}_{t'}(x)$,
with $\hat{C}_{\inf(\mathcal{T})} = \emptyset$ and $\hat{C}_{\sup(\mathcal{T})} = \mathbb{R}$.
Then, prediction intervals $\hat{C}^{\text{NCP}}(x) = \hat{C}_{\tau}(x)$, where
\vspace{-0.4cm}
{\setlength{\abovedisplayskip}{0pt}
\setlength{\belowdisplayskip}{2pt} 
  
\begin{equation}
\begin{aligned}
    \label{eq:NCP:conformalization}
    \tau = \inf_{t \in \mathcal{T}} \left\{
        t \, \middle| \,
        \sum_{i \in \mathcal{I}_{\text{cal}}} \mathbf{1}_{\hat{C}_{t}(X_{i})}(Y_{i}) \geq (1 {-} \alpha)(1 {+} |\mathcal{I}_{cal}|)
    \right\}
\end{aligned}
\end{equation}} \vspace{-0.1cm} \\
is computed from the calibration set.
This framework encompasses all types of conformity scores, while maintaining the theoretical guarantees of standard CP \cite{gupta2022nested}.
For instance, the CQR construction can be cast in the NCP framework by considering
{
  \setlength{\abovedisplayskip}{3pt}
  \setlength{\belowdisplayskip}{5pt}
\begin{align}
    \label{eq:NCP:CQR_nested}
    \hat{C}_{t}(x) = \left[
        \hat{q}_{\frac{\alpha}{2}}(x) - t, \ \hat{q}_{1-\frac{\alpha}{2}}(x) + t
    \right].
\end{align}}

To ensure clarity, the remainder of the paper is presented using the NCP framework.
Namely, each CP methodology is presented by stating the corresponding family of nested prediction sets;
prediction intervals are then constructed using the NCP approach and \eqref{eq:NCP:conformalization}.

\paragraph{Other Related Methods} 
CP techniques have been used to analyze the precision of sketching algorithms, which are increasingly vital tools for handling massive datasets in modern machine learning \cite{broder2004network, goyal2012sketch, cormode2018privacy, zhang2014these}.
\citet{sesia2023conformal} construct prediction intervals 
for the frequency of a queried object based on a sketch \cite{charikar2002finding}. 
Therein, a deterministic upper bound \(\hat{B}^u\) provided by
\cite{cormode2005improved} is combined with a trivial lower bound
\(\hat{B}^l = 0\) (reflecting the non-negativity of counts) to construct
nested intervals. The construction, referred to as ``SFD CP'' in
this paper, is defined as:
\vspace{-0.5cm}
\begin{align*} 
    \label{sesia's}
    \hat{C}^{\text{SFD}}_t(x) = \big[\max\{0, \hat{B}^u(x) {-} t\}, \min\{\hat{B}^u(x), t\}\big].
\end{align*}
\citet{sesia2023conformal} also introduces an adaptive version of this
construction, designed to better account for heteroscedasticity in
residuals. The CPUL framework generalizes this approach by
combining information from $\Bl, \Bu$ in multiple ways; see Section \ref{sec:CPUL:CPUL}.

Another related line of work leverages information from multiple
models fitted on the training set. For example,
\cite{liang2024conformal, yang2024selection} focus on selecting the
model that produces the most efficient prediction intervals. Extending
these ideas, the CPUL framework reinterprets the problem in the
context of multiple models, simultaneously utilizing both upper and
lower bounds to maximize the utility of the available information.

\section{{Methodology}}
\setlength{\textfloatsep}{5pt} 
\begin{figure}[!t]
    \centering
    \begin{subfigure}[b]{0.48\columnwidth}
        \centering
\includegraphics[width=\linewidth]{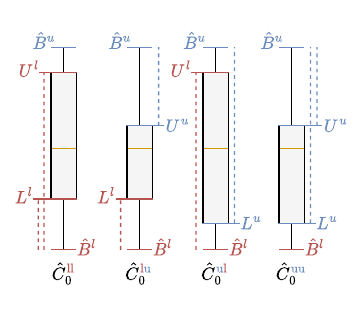}
        \caption{Illustration of the construction of prediction intervals in CPUL: (a) $\Cll_{0}$, (b) $\Cuu_{0}$, (c) $\Clu_{0}$, (d) $\Cul_{0}$. Each family of prediction intervals is conformalized following the NCP framework (see  ~\eqref{eq:NCP:conformalization}).}
        \label{fig:CPUL}
    \end{subfigure}
    \hfill
    \begin{subfigure}[b]{0.48\columnwidth}
        \centering
        \includegraphics[width=\linewidth]{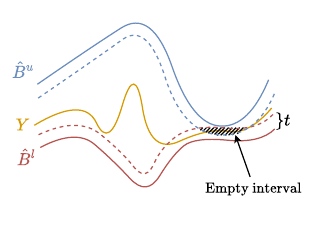}
        \caption{Illustration of Paradoxical Miscoverage (motivation for OMLT): using a constant offset ($\pm t$) in the NCP construction results in an empty prediction interval where $\Bu(x) - \Bl(x)$ is small.}
        \label{fig:Paradoxical_Miscoverage}
    \end{subfigure}
    \caption{Illustration of CPUL-OMLT construction}
\end{figure}
This section presents CPUL-OMLT, a CP framework specifically designed to exploit the favorable properties of $\Bl$ and $\Bu$. Contrary to the general CP setting, $\Bl$ and $\Bu$ are not point estimates; they yield valid bounds on the target variable, a setting overlooked by traditional CP methods like SCP and CQR. Additionally, standard CP methods do not necessarily integrate information from both bounds, nor are they inherently designed to account for heteroskedastic residuals between these bounds (e.g., $\Bl$ might provide accurate estimates while $\Bu$ performs poorly). The results in Section \ref{sec:experiments} further illustrate these observations.
\vspace{-0.3cm}
\subsection{Exploiting valid bounds in NCP}
\label{sec:CPUL:strenghtening}
\vspace*{-4pt}
Recall that $\Bl, \Bu$ provide valid lower and upper bounds on the target variable, i.e., $\Bl(X) \, {\leq} \, Y \, {\leq} \, \Bu(X)$ always holds.
This suggests a simple procedure for strengthening $\hat{C}$ without loss of coverage (as shown by Proposition \ref{thm:CPUL:strengthening}) into
{  \setlength{\abovedisplayskip}{3pt}
  \setlength{\belowdisplayskip}{0pt}
  \begin{equation}\begin{aligned}
    \label{eq:strengthening}
    \tilde{C}(x) = \hat{C}(x) \cap [\Bl(x), \Bu(x)].
\end{aligned}\end{equation}}
\vspace{-0.4cm}
\begin{propositionrep}
\label{thm:CPUL:strengthening}
    Let $\hat{C}(\cdot)$ denote a prediction interval with coverage $1
    \,{-}\, \alpha$, i.e.,
    {$\mathbb{P} \left(Y_{N+1} \, {\in} \, \hat{C}(X_{N+1}) \right) \, {=} \, 1 \, {-} \, \alpha$} for some $\alpha \, {\in} \, [0, 1]$.
    Next, define the strengthened interval $\tilde{C}(x) \, {:=} \, $ $\hat{C}(x) \, {\cap} \, \big[\Bl(x), \Bu(x)\big], \forall x \in \mathcal{X}$.
    Then, {$\mathbb{P} \big(Y_{N+1} \, {\in} \, \tilde{C}(X_{N+1})\big)
    =\mathbb{P} \big(Y_{N+1} \, {\in} \, \hat{C}(X_{N+1}) \big)
        = 1 {-} \alpha.$}
\end{propositionrep}
\vspace{-5pt}
\begin{appendixproof}
    It suffices to prove that {$\mathbb{P} \left(Y \in \tilde{C}(X)\right) = \mathbb{P} \left(Y \in \hat{C}(X) \right)$}.
    First note that 
    \begin{align*}
        \hat{C}(X) &= \hat{C}(X) \cap \mathbb{R}\\
        &= \hat{C}(X) \cap \left( 
            (-\infty, \Bl(X)) \cup [\Bl(X), \Bu(X))] \cup (\Bu(X), +\infty)
        \right)
    \end{align*}
    and note that $(-\infty, \Bl(X))$, $[\Bl(X), \Bu(X))]$ and $(\Bu(X), +\infty)$ are disjoint.
    Also note that
    \begin{align*}
        \mathbb{P} \left(Y \in \hat{C}(X) \cap (-\infty, \Bl(X))\right) 
        = 
        \mathbb{P} \left(Y \in \hat{C}(X) \cap (\Bu(X), +\infty)\right) 
        = 
        0,
    \end{align*}
    because $\Bl(X) \leq Y \leq \Bu(X)$ by definition of $\Bl, \Bu$.
    It then follows that
    {
    \begin{align*}
        \mathbb{P} \left(Y \in \hat{C}(X)\right)
            &= \mathbb{P} \left(Y \in \hat{C}(X) \cap (-\infty, \Bl(X))\right)
            + \mathbb{P} \left(Y \in \hat{C}(X) \cap [\Bl(X), \Bu(X))]\right)\\
            & + \mathbb{P} \left(Y \in \hat{C}(X) \cap (\Bu(X), +\infty)\right)\\
            &= 0 + \mathbb{P} \left(Y \in \hat{C}(X) \cap [\Bl(X), \Bu(X))]\right) + 0\\
            &= \mathbb{P} \left(Y \in \tilde{C}(X)\right)
    \end{align*}}
    which concludes the proof.
\end{appendixproof}
It is important to note that Proposition \ref{thm:CPUL:strengthening} holds irrespective of how the original prediction interval $\hat{C}$ is obtained.
Theorem \ref{thm:CPUL:nested_strengthening} shows that, if the calibration step is performed using the NCP framework, then there is no loss of performance whether the strengthening is performed before or after calibration.
\begin{theoremrep}
\label{thm:CPUL:nested_strengthening}
    Consider a family of nested prediction sets $\{\hat{C}_{t}\}_{t \in \mathcal{T}}$, where $\mathcal{T} \, {\subseteq} \, \mathbb{R}$, and let $\hat{\tau}$ be obtained following the NCP calibration step as per  \eqref{eq:NCP:conformalization}.
    Next, define the family of nested strengthened intervals $\{\tilde{C}_{t}\}_{t \in \mathcal{T}}$, where
    $
        \forall x \in \mathbb{R}^{d}, \forall t \in \mathcal{T}, \tilde{C}_{t}(x) = \hat{C}_{t}(x) \cap \big[ \Bl(x), \Bu(x) \big],$
    and let $\tilde{\tau}$ be obtained from  \eqref{eq:NCP:conformalization}.
    Then, $\hat{\tau} = \tilde{\tau}$ and $\forall x \in \mathbb{R}^{d},$
    $\hat{C}_{\hat{\tau}}(x) \cap [\Bl(x), \Bu(x)]
        = \tilde{C}_{\tilde{\tau}}(x)$.
\end{theoremrep}
\begin{appendixproof}
    First note that $\{\tilde{C}_{t}\}_{t \in \mathcal{T}}$ is indeed a family of nested intervals that satisfies the NCP assumptions; this follows from the fact that $\tilde{C}_{t} \subseteq \hat{C}_{t}, \forall t$.

    Next, using the same argument as the proof of Proposition \ref{thm:CPUL:strengthening}, 
    \begin{align}
        \forall i \in \mathcal{I}_{\text{cal}}, \mathbf{1}_{\hat{C}_{t}(X_{i})}(Y_{i}) = \mathbf{1}_{\tilde{C}_{t}(X_{i})}(Y_{i})
    \end{align}
    Substituting this in  \eqref{eq:NCP:conformalization} then yields
    \begin{align}
        \left\{
            t 
        \, \middle| \,
            \sum_{i \in \mathcal{I}_{\text{cal}}} \mathbf{1}_{\tilde{C}_{t}(X_{i})}(Y_{i}) \geq (1 {-} \alpha)(1 {+} |\mathcal{I}_{cal}|)
        \right\}
        &= 
        \left\{
            t 
        \, \middle| \,
            \sum_{i \in \mathcal{I}_{\text{cal}}} \mathbf{1}_{\hat{C}_{t}(X_{i})}(Y_{i}) \geq (1 {-} \alpha)(1 {+} |\mathcal{I}_{cal}|)
        \right\},
    \end{align}
    which then yields $\hat{\tau} = \tilde{\tau}$ by unicity of the infimum, and $\hat{C}_{\hat{\tau}}(X) \cap [\Bl(X), \Bu(X)] = \tilde{C}_{\tilde{\tau}}(X)$ follows immediately. \end{appendixproof}
\vspace{-0.2cm}
Theorem \ref{thm:CPUL:nested_strengthening} allows decoupling of the calibration step from the strengthening.
Thereby, one can perform the calibration step without access to $\Bl$ or $\Bu$, without any loss of coverage.
This is particularly valuable when $\Bl$ or $\Bu$ are not available during training/calibration due to, e.g., privacy concerns.
\vspace*{-6pt}
\subsection{Conformal Prediction from upper and lower bound models (CPUL)}
\vspace*{-5pt}
\label{sec:CPUL:CPUL}
The proposed CPUL framework combines several steps to fully exploit the knowledge that $\Bl$ and $\Bu$ provide valid bounds on the target variable, and to account for the heteroskedasticity of their residuals.
The algorithm is presented in the NCP framework and, for ease of reading, the presentation omits the strengthening of prediction intervals described in  \eqref{eq:strengthening}, i.e., strengthening is always performed implicitly.
Recall that, by Theorem \ref{thm:CPUL:nested_strengthening}, this does not affect the conformalization procedure.

First define the residuals {\setlength{\abovedisplayskip}{1pt}
  \setlength{\belowdisplayskip}{0pt}$\rl \, {=} \, Y \, {-} \, \Bl(X)$} and {$\ru \, {=} \, Y \, {-} \, \Bu(X)$}, and denote by {$\hat{Q}^{l}_{\beta}$, $\hat{Q}^{u}_{\beta}$} the {$\beta$} quantiles of {$\rl, \ru$}, evaluated on the training set. Then define {\setlength{\abovedisplayskip}{3pt}
  \setlength{\belowdisplayskip}{0pt} \begin{subequations}
\label{eq:CPUL:bounds}
\begin{align}
    \label{eq:CPUL:lb:l}
    \Ll(x) &= \Bl(x) + \hat{Q}^{l}_{\frac{\alpha}{2}},\\
    \label{eq:CPUL:lb:u}
    \Lu(x) &= \Bu(x) + \hat{Q}^{u}_{\frac{\alpha}{2}},\\
    \label{eq:CPUL:ub:l}
    \Ul(x) &= \Bl(x) + \hat{Q}^{l}_{1-\frac{\alpha}{2}},\\
    \label{eq:CPUL:ub:u}
    \Uu(x) &= \Bu(x) + \hat{Q}^{u}_{1-\frac{\alpha}{2}}.
\end{align}
\end{subequations}}
Note that $\Ll$, $\Lu$, $\Ul$, $\Uu$ do not provide valid lower nor upper bounds on the target variable ($Y$).
The motivation for adjusting the initial predictors $\Bl, \Bu$ using quantiles $\hat{Q}^{l}, \hat{Q}^{u}$ is to account for the heteroskedascitity of $\rl, \ru$.
Note that the construction in  \eqref{eq:CPUL:bounds} only requires evaluating the residuals $\rl, \ru$ on the training set, and extracting their quantiles. 

Next, $\Ll$, $\Lu$, $\Ul$, $\Uu$ are combined to form four families of nested prediction intervals $\{\Cll_{t}\}_{t \in \mathbb{R}}$, $\{\Clu_{t}\}_{t \in \mathbb{R}}$, $\{\Cul_{t}\}_{t \in \mathbb{R}}$, $\{\Cuu_{t}\}_{t \in \mathbb{R}}$ as follows:\\ 
\vspace{-0.3cm}
{\setlength{\abovedisplayskip}{0pt}
\setlength{\belowdisplayskip}{3pt} \begin{subequations}
\label{eq:CPUL:nested}
\begin{align}
    \label{eq:CPUL:ll}
    \Cll_{t}(x) &= \big[\Ll(x) - t, \Ul(x) + t \big],\\
    \label{eq:CPUL:lu}
    \Clu_{t}(x) &= \big[\Ll(x) - t, \Uu(x) + t \big],\\
    \label{eq:CPUL:ul}
    \Cul_{t}(x) &= \big[\Lu(x) - t, \Ul(x) + t \big],\\
    \label{eq:CPUL:uu}
    \Cuu_{t}(x) &= \big[\Lu(x) - t, \Uu(x) + t \big].
\end{align}
\end{subequations}}
This construction is illustrated in Figure \ref{fig:CPUL}.
Each family is then conformalized using the NCP procedure on the calibration set, and the model with smallest width is selected.

Algorithm \ref{alg:CPUL} summarizes the proposed CPUL method, which proceeds as follows.
First, the dataset $\mathcal{D}$ is split into training and calibration sets (line 1).
Then, lower and upper bound predictors (i.e., $\Bl, \Bu$) are fit using the training set (line 2), 
after which quantiles of the corresponding residuals are evaluated on the training set (line 3).
Next, four families of nested intervals are constructed following  \eqref{eq:CPUL:nested},
each of which is calibrated using the NCP framework using \eqref{eq:NCP:conformalization} with the calibration set $\mathcal{D}_{\text{cal}}$ (line 4).
The final step of the algorithm (line 5) selects, among the four models $\Cll, \Clu, \Cul, \Cuu$, the one with smallest average width over the calibration set (as an estimator of $\mathbb{E} \big| \hat{C}(X_{N+1}) \big|$).
It is important to note that the model selection step is not performed on a per-sample basis.
Rather, Algorithm 1 selects \emph{one} variant among $\Cll, \Clu, \Cul, \Cuu$, which is then used across the entire test set.
For instance, if {$\Clu_{\tlu}$ is selected, then $\hat{C}^{**} \, {=} \, \Clu_{\tlu}$} and the CPUL prediction interval is
$\hat{C}^{**}(X_{N+1}) \, {=} \, \Clu_{\tlu}(X_{N+1})$.
Theorem \ref{thm:CPUL:coverage} provides theoretical guarantees on the coverage of $\hat{C}^{**}$.
\begin{theoremrep}
\label{thm:CPUL:coverage} 
Assume that $\{(X_{1}, Y_{1}), ..., (X_{N+1}, Y_{N+1})\}$ are i.i.d. samples, and let $\hat{C}^{**}$ be the CPUL model selected by Algorithm \ref{alg:CPUL}. {Define $N_{\text{cal}} = |\mathcal{D}_{\text{cal}}|$ and $\eta = \sqrt{\nicefrac{\log(8)}{2}} + \nicefrac{1}{3}$. } 
Then 
{\setlength{\abovedisplayskip}{5pt}
  \setlength{\belowdisplayskip}{0pt} 
{
\begin{align*}
\label{eq:CPUL:coverage_guarantee_generalization}
    \mathbb{P} \left( Y_{N+1} \in \hat{C}^{**}(X_{N+1}) \right) 
    \geq 
    \frac{1 {+} N_{\text{cal}}}{N_{\text{cal}}} (1 {-} \alpha) - \frac{\eta}{\sqrt{N_{\text{cal}}}}.
\end{align*}}}
\end{theoremrep}
\vspace{-0.5cm}
\begin{appendixproof}
    The result follows directly from Theorem 1 from \citet{yang2024selection}.
\end{appendixproof} 

\subsection{Relation to other methods}
\label{sec:CPUL:other_methods}
Several parallels can be drawn between CPUL and existing CP methodologies.
For instance, the use of residual quantiles $\hat{Q}^{l}, \hat{Q}^{u}$, when constructing the nested prediction sets $\Cll, \Cuu$, is related to the Split CP method.
Indeed, applying SCP to $\Bl$ or $\Bu$ yields conformal prediction intervals (see \eqref{eq:SCP:predict_interval}) that closely resemble the structure of $\Cll$ and $\Cuu$.
Therefore, given that CPUL also exploits the $\Clu$ and $\Cul$ variants, one should expect CPUL to consistently outperform SCP.

The $\Cul$ construction is also related to the adaptive approach followed in \citet{sesia2023conformal}.
The main difference between the two is the use of empirical quantiles $\hat{Q}^{l}, \hat{Q}^{u}$ in CPUL, which can be obtained much more efficiently, compared to training additional quantile regression models as in \citet{sesia2023conformal}.
The latter approach provides finer local adaptivity, albeit at a higher computational cost. 
\begin{algorithm}[ht]
\caption{CPUL}
\label{alg:CPUL}
    \mbox{\textbf{Input:} $\mathcal{D} \, {=} \, ({X_i},  {Y_i})_{i=1}^N$, $\alpha \, {\in} \, [0, 1]$.}
    \mbox{\textbf{Output:} Selected model $\hat{C}^{**}$.} 
\begin{algorithmic}[1]
    \STATE \label{alg:CPUL:data_split}
        Split $\mathcal{D}$ into training and calibration sets $\mathcal{D}_{\text{train}}$ and $\mathcal{D}_{\text{cal}}$
    \STATE \label{alg:CPUL:train}
        Train lower/upper bound models $\Bl, \Bu$ using $\mathcal{D}_{\text{train}}$
    \STATE \label{alg:CPUL:quantiles}
        Compute empirical quantiles $\hat{Q}^{l}_{\frac{\alpha}{2}}$, $\hat{Q}^{l}_{1-\frac{\alpha}{2}}$, $\hat{Q}^{u}_{\frac{\alpha}{2}}$, $\hat{Q}^{u}_{1-\frac{\alpha}{2}}$ on the training set $\mathcal{D}_{\text{train}}$
    \STATE \label{alg:CPUL:calibration}
        Form $\{\Cll_{t}\}_{t \in \mathbb{R}}$, $\{\Clu_{t}\}_{t \in \mathbb{R}}$, $\{\Cul_{t}\}_{t \in \mathbb{R}}$, $\{\Cuu_{t}\}_{t \in \mathbb{R}}$ as per  \eqref{eq:CPUL:nested}, and compute $\tll$, $\tlu$, $\tul$, $\tuu$ using  \eqref{eq:NCP:conformalization}.
    \STATE \label{alg:CPUL:selection}
        Select model with smallest width on calibration set
      {\setlength{\abovedisplayskip}{0pt}
  \setlength{\belowdisplayskip}{0pt}\begin{align*}
            \hat{C}^{**} :=
            \underset{\hat{C} \in \{ \Cll_{\tll}, \Clu_{\tlu}, \Cul_{\tul}, \Cuu_{\tuu} \}}{\text{argmin}}
            \frac{1}{|\mathcal{D}_{\text{cal}}|}
            \sum_{X_{i} \in \mathcal{D}_{\text{cal}}}
            \left|\hat{C}(X_i)\right| 
        \end{align*}}
\end{algorithmic}
\end{algorithm}
\setlength{\textfloatsep}{8pt}

\subsection{Conformal Prediction from upper and lower bound models with Optimal Minimal Length 
Threshold (CPUL-OMLT)}\label{sec:OMLT}

A key limitation of CPUL {alone}, which it shares with most CP methods, is its tendency to produce overly narrow prediction intervals, particularly {where} the initial bounds \(\Bu(x)\) and \(\Bl(x)\) are tight. 
The constant adjustment of $t$ in the NCP framework becomes disproportionately large, in regions where the bounds are extremely tight, leading to empty prediction intervals and, in turn, under-coverage in such regions.
Figure \ref{fig:Paradoxical_Miscoverage} illustrates this issue in the CPUL-lu setting. 
After calibrating the initial bounds, the area around $x_0$ remains uncovered due to the empty intersection of the calibrated intervals.
This behavior results in inefficient prediction intervals, as the method should over-cover the less confident regions (i.e., areas where \(\Bu(x) - \Bl(x)\) is large). 
Hence, paradoxically, the regions with tightest initial bounds become most vulnerable and become under-covered.
\cite{sesia2020comparison} proposed CQR-r, with
{ $
\hat{C}_t^{\text{CQR-r}}(x) = \left[\hat{L}_t^{\text{CQR-r}}(x), \hat{U}_t^{\text{CQR-r}}(x)\right] \ \text{for } x \in \mathcal{X}, $} where 
{\setlength{\abovedisplayskip}{5pt}
  \setlength{\belowdisplayskip}{5pt} {\begin{align*}
\hat{L}_t^{\text{CQR-r}}(x) =& \hat{q}_{\alpha/2}(x) - t \left(\hat{q}_{1-\alpha/2}(x) - \hat{q}_{\alpha/2}(x)\right); \\ 
\hat{U}_t^{\text{CQR-r}}(x) =& \hat{q}_{1-\alpha/2}(x) + t \left(\hat{q}_{1-\alpha/2}(x) - \hat{q}_{\alpha/2}(x)\right).
\end{align*}}}

The design of \cite{sesia2020comparison} scales \(t\) by 
\(\hat{q}_{1-\alpha/2}(x) - \hat{q}_{\alpha/2}(x)\), which mitigates the disproportionate reduction in prediction interval size. However, as noted in \cite{sesia2020comparison}, its overall efficiency is worse than the original fixed-length version \eqref{eq:NCP:CQR_nested}. Thus, it remains unclear whether such scaling, despite being adaptive to the initial tightness of the base models, is preferred over the fixed-length adjustment \(t\). The experiments in section \ref{sec:experiments} include a CQR-r adaptation tailored to the setting of this paper for comparison.

To address this paradoxical miscoverage, the paper proposes the
optimal minimal length threshold method (OMLT) as an alternative to the
relative scaling approach used in \citet{sesia2020comparison}.
The core idea of OMLT consists in introducing a threshold $\ell \geq 0$,
representing the minimum allowed length for a prediction interval. OMLT identifies regions where the given $\hat{B}^u$ and $\hat{B}^l$ are tightest, where even minor shrinking during calibration poses a high risk of significant undercoverage below the desired level of $1 - \alpha$. The design of OMLT retains the standard fixed-length adjustment of $t$ and introduces a threshold for the minimal allowed prediction interval length, effectively marking the boundary of high-risk regions.

Consider a family of nested intervals $\{\hat{C}_{t}\}_{t \in \mathcal{T}}$ satisfying the NCP assumptions,
and define{\setlength{\abovedisplayskip}{5pt}
  \setlength{\belowdisplayskip}{5pt} 
\begin{align*}
    \kappa_{\ell}(x) = \inf_{t \in \mathcal{T}} \big\{ t \ \big| \ \ell \leq |\hat{C}_{t}(x)| \big\},
\end{align*}}
from which a new family $\{\bar{C}_{t}\}_{t \in \mathcal{T}}$ is constructed as 
{\setlength{\abovedisplayskip}{3pt}
  \setlength{\belowdisplayskip}{3pt} {
\begin{align*}
    \bar{C}_{\ell,t}(x) = 
    \left\{
    \begin{array}{ll}
         \hat{C}_{t}(x)         & \text{if } (\Delta(x) \, {\geq}  \,\ell) \wedge (t \, {>} \, \kappa_{\ell}(x))\\
         \hat{C}_{\kappa_{\ell}(x)}(x) & \text{if } (\Delta(x) \, {\geq}  \,\ell) \wedge (t \, {\leq} \, \kappa_{\ell}(x)) \\
         {[\Bl(x), \Bu(x)]}     & \text{if } (\Delta(x) \, {\leq}  \,\ell)
    \end{array}
    \right.
\end{align*}}}%
where $\Delta(x) \, {=} \, \Bu(x) \, {-} \, \Bl(x)$.
It is easy to verify that constructed family of intervals $\{\bar{C}_{\ell,t}\}_{t \in \mathcal{T}}$ is a nested family, therefore satisfying the coverage guarantee given in \eqref{eq:NCP:conformalization}.
The optimal minimum length threshold can be obtained as the solution of the optimization problem \eqref{eqn:OMLT}, though an exact solution is not needed.{\setlength{\abovedisplayskip}{3pt}
  \setlength{\belowdisplayskip}{3pt} 
 \begin{subequations}
\label{eqn:OMLT}
{\begin{align}
\min_{\ell \geq 0, t \in \mathcal{T}} \quad & \mathbb{E}_X \left[ \left| \bar{C}_{\ell,t}(X) \right|  \right] \\
\text{s.t.} \quad & \mathbb{P} \left( f(X) \in \bar{C}_{ \ell,t}(X)  \right) \geq 1 - \alpha.
\end{align}}
\end{subequations}} 
The key idea underlying OMLT is that the size of the prediction interval should not be smaller than \(\ell\), which prevents under-coverage when a prediction interval is too small. The only exception is for the tightest regions, i.e., when \(\hat{B}^u(x) - \hat{B}^l(x) \leq \ell\), in which case there is no need to enlarge \(\bar{C}_{\ell,t}(x)\) beyond \(\ell\). The original upper and lower bounds can be confidently relied upon, as \([\hat{B}^l(x), \hat{B}^u(x)]\) already provides high-quality coverage.

Since \eqref{eqn:OMLT} reduces to the original problem in \eqref{eq:prediction_interval} when \(\ell = 0\) for arbitrary $\{\hat{C}_t\}_{t\in\mathcal{T}}$, the formulation of CPUL-OMLT is designed to produce prediction intervals whose expected length is not always greater than that of CPUL, at the same \(1 - \alpha\) coverage. Moreover, OMLT has the potential to extend beyond this setting and reduce the average prediction interval length in cases where disproportionate prediction calibration is observed. For more details, see Section \ref{sec:CPUL and OMLT implementation details}. 

\section{Experiments}\label{sec:experiments}
The performance of CPUL and CPUL-OMLT are evaluated to demonstrate
their ability to achieve valid coverage while producing narrower
prediction intervals. The experiments focus on UQ for the optimal value of  economic dispatch
problems, detailed in Appendix \ref{sec:background:optimization}.  
UQ is conducted given lower and upper bounds derived from
primal-dual optimization proxy models \cite{chen2023end, qiu2024dual,
  klamkin2024dual, chen2024real, klamkin2025self}.
Namely, the dual proxy provides valid lower bounds ($\Bl$), and the primal proxy provides valid upper bounds $\Bu$). Further details on these optimization proxies are provided in Appendices \ref{sec:background:optimization}
and \ref{sec:background:proxies}.
\begin{table}[!t]
    \centering
    \setlength{\tabcolsep}{4pt} 
    \begin{threeparttable}
    \caption{Performance Comparison of CP Methods ($\alpha = 10\%$)} 
    \label{table:results}
    \scriptsize 
    \begin{tabular}{@{}l|cc|cc|cc@{}} 
       \toprule
        & \multicolumn{2}{c}{\textbf{89\_pegase}} & \multicolumn{2}{c}{\textbf{118\_ieee}} & \multicolumn{2}{c}{\textbf{1354\_pegase}} \\
        \cmidrule(lr){2-3} \cmidrule(lr){4-5} \cmidrule(lr){6-7}
     
        \textbf{UQ Method} & \textbf{PICP (\%)} & \textbf{Size (\%)} & \textbf{PICP (\%)} & \textbf{Size (\%)} & \textbf{PICP (\%)} & \textbf{Size (\%)} \\
   
        \midrule 
        $[\Bl, \hat{B^u}]$ & 100.0 (0.00) & 0.410 (0.016) & 100.0 (0.00) & 0.281 (0.085) & 100.0 (0.00) & 3.949 (3.807) \\
        \midrule 
        Split CP w/ $\Bl$ & 90.15 (0.60) & 0.197 (0.007) & 90.34 (0.63) & \color{red}{0.206 (0.146)} & 89.60 (0.54) & \color{blue}{1.308 (1.160)} \\
        Split CP w/ $\Bu$ & 90.02 (0.54) & \color{red}{0.205 (0.007)} & 90.00 (0.50) & \color{blue}{0.105 (0.004)} & 89.40 (0.49) & \color{red}{1.570 (0.853)} \\
        SFD CP             & 91.23 (0.56) & \color{blue}{0.188 (0.007)} & 90.10 (0.50) & 0.135 (0.013)          & 90.22 (0.27) & 1.456 (0.609)          \\
        CQR                & 91.40 (1.80) & \color{red}{0.389 (0.018)} & 90.08 (2.75) & \color{red}{0.190 (0.147)} & 90.13 (0.43) & \color{red}{3.501 (3.939)} \\
        CQR-r              & 90.71 (0.72) & \color{red}{0.332 (0.020)} & 90.88 (1.33) & \color{red}{0.180 (0.167)} & 91.57 (1.32) & \color{red}{3.271 (3.997)} \\
        CPUL {(ours)}               & 91.23 (0.56) & \color{blue}{0.188 (0.007)} & 90.02 (0.46) & \color{blue}{0.105 (0.004)} & 89.65 (0.56) & \color{blue}{1.306 (1.162)} \\
        CPUL-OMLT {(ours)}          & 90.28 (0.51) & \color{blue}{0.187 (0.008)} & 90.01 (0.48) & \color{blue}{0.103 (0.007)} & 89.69 (0.49) & \color{blue}{1.037 (0.890)} \\

        \bottomrule
    \end{tabular}
    \begin{tablenotes}

        \item[*] \scriptsize For each dataset, the three shortest intervals are colored blue, while the three largest intervals are colored red.
    \end{tablenotes}
    \end{threeparttable}
\end{table}
\setlength{\textfloatsep}{0pt}
\paragraph{Datasets} Experiments are performed over three datasets: 118\_ieee~\cite{ieee_pstca}, 1354\_pegase~\cite{pegase}, and 89\_pegase~\cite{pegase}, corresponding to power grids of different sizes. For each dataset, samples are randomly shuffled and split 10 times. Dataset details are provided in Section \ref{data123}. Evaluation metrics reported are the mean and standard deviation across these 10 runs. 
\paragraph{Baselines}  
The performance of CPUL-OMLT is compared against several CP baselines.  Recall that, unless specified otherwise, all prediction intervals are strengthened as per  \eqref{eq:strengthening}.
The CP baselines include: Split CP (\ref{eq:SCP:predict_interval}), using either $\Bl$ or $\Bu$ as base predictors; CQR and its variant CQR-r, wherein $\Bl$ and $\Bu$ are treated as the initial lower and upper quantile regressors in the CQR construction;
an adapted SFP CP approach of \citet{sesia2023conformal}.
Note that the latter matches the $\Cul$ construction.
Additional implementation details are provided in Appendix \ref{app:experiments:cp_baselines}.
\paragraph{Evaluation Metrics}
All methods are evaluated on a held out test set $\mathcal{D}_{\text{test}} \, {=} \, (X_{i}, Y_{i})_{i \in \mathcal{I}_{\text{test}}}$, using two evaluation criteria: \emph{coverage} and \emph{interval length}.
The former is evaluated via the \emph{Prediction Interval Coverage Percentage} (PICP), which measures the proportion of true values contained within the predicted intervals on the test set, i.e., $
\text{PICP} = \frac{1}{|\mathcal{I}_{\text{test}}|} \sum_{i\in \mathcal{I}_{\text{test}}} \mathbf{1}_{\hat{C}(X_i)}(Y_{i}).$ 

At a given confidence level \(1 \, {-} \, \alpha\),
shorter interval lengths are considered more desirable. 
This is captured through \emph{expected normalized length} of prediction interval $\hat{C}$, defined as $
    \hat{E}_X(\hat{C}) 
    = 
    |\mathcal{I}_{\text{test}}|^{-1}
    \sum_{i \in \mathcal{I}_{\text{test}}} \big(
        \left|Y_i\right|^{-1} |\hat{C}(X_i)| \big)$,
where $Y_{i}$ and $\hat{C}(X_i)$ denote the (true) optimal value and the prediction interval for sample $i$.
The paper considers the scaled interval length $|\hat{C}(X_{i})| / |Y_{i}|$, expressed as a percentage, rather than absolute interval length $|\hat{C}(X_{i})|$, to account for the possibly large range of values taken by $Y$.
This metric is routinely used in the optimization literature  \cite{chen2023end, vivas2020systematic, wang2009comparison}, where it is referred to as \emph{optimality gap}.
All metrics are reported as the mean and standard deviation over 10 runs.
\paragraph{Experiment Results}
\setlength{\textfloatsep}{5pt} 
\begin{figure*}
    \centering    \includegraphics[width=0.99\linewidth]{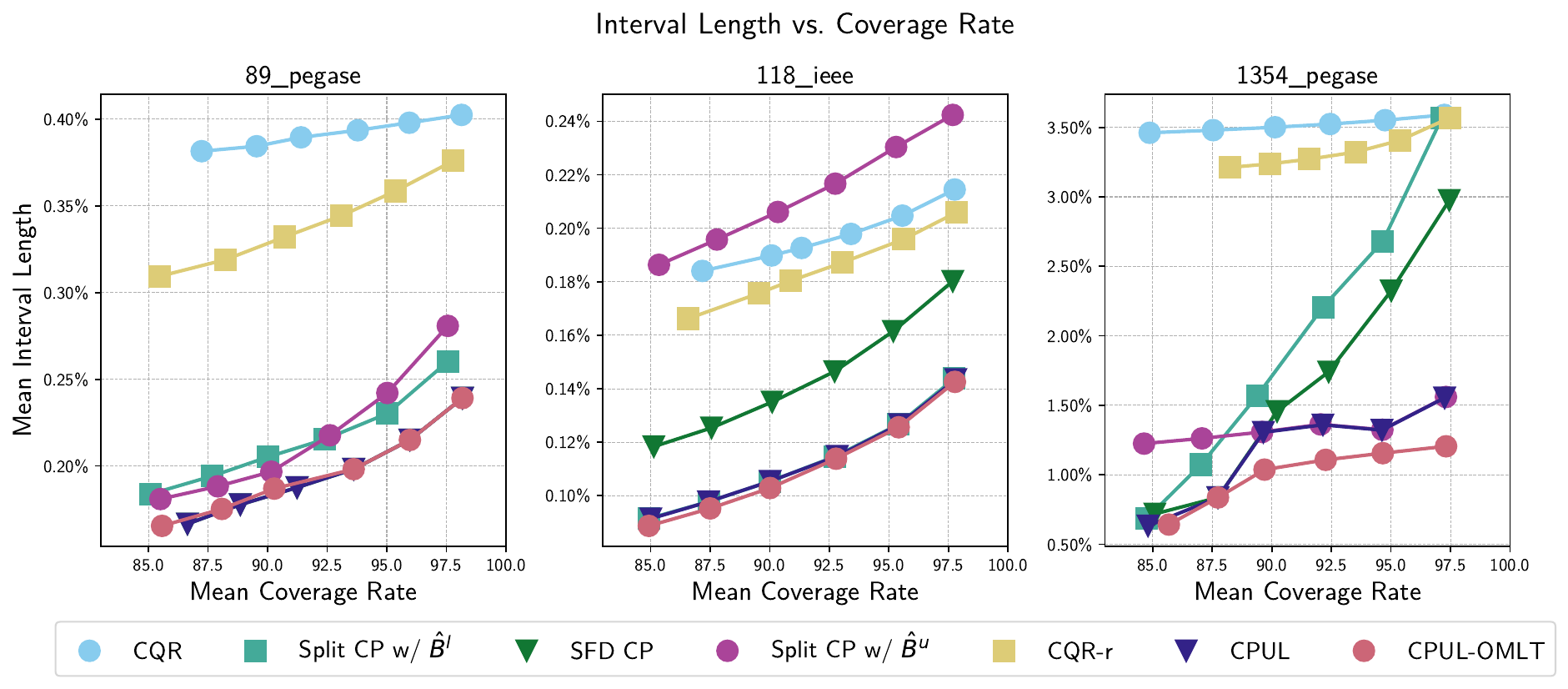}
    \caption{Performance comparison across datasets (Note: in 89\_pegase, SFD CP overlaps with CPUL and in 118\_ieee, Split CP w/ $\Bl$  overlaps with CPUL.)}  \label{fig:cp_comparison}
\end{figure*}
Table \ref{table:results} and Figure \ref{fig:cp_comparison} present the numerical performance of the various methods.
Table \ref{table:results} presents, for every dataset, the average and standard deviation of each method's interval length and coverage, both expressed as a percentage.
The table reports results for $\alpha=10\%$; additional $\alpha$ values are reported in Table \ref{table full:results} in Appendix \ref{sec:Additional experiment results}.
Figure \ref{fig:cp_comparison} displays the interaction between each method's interval length and coverage, across a broader range of $\alpha$ values.
This provides a more global view of each method's overall performance.

The results in Table \ref{table:results} demonstrate that all CP methods achieve satisfactory coverage levels, i.e., test coverage is typically close to $1 - \alpha$, which is expected given that all methods are properly calibrated.
Moreover, all prediction intervals yield a substantial reduction in size compared to the original prediction $[\Bl, \Bu]$.
For instance, a reduction of only 10\% in coverage (i.e., 90\% prediction intervals) can yield up to a two- and three-fold reduction in interval size.
This demonstrates the value of using UQ techniques to provide more actionable information to practitioners.

Furthermore, CPUL and CPUL-OMLT consistently achieve state-of-the-art performance, across datasets and target coverage.
On the other hand, CQR and CQR-r exhibit the worst performance overall, likely due to their inability to properly
address heteroskedasticity.
This is most evident in Figure \ref{fig:cp_comparison}.
On the 1354\_pegase dataset (which corresponds to a larger power grid) CPUL-OMLT achieves the smallest interval size. In particular, CPUL-OMLT produces prediction intervals whose width is about 10\% smaller than the second-best method, CPUL.
These results highlight the important of the model selection procedure in CPUL.

Figure \ref{fig:cp_comparison} demonstrates that, while methods
like Split CP and SFD-CP deliver good
results occasionally, their performance is highly variable across different
datasets and confidence levels.
For instance, although SFD-CP is among the best-performing methods on 89\_pegase, its performance on 1354\_pegase is significantly worse than CPUL when coverage is close to 100\% (corresponding to small values of $\alpha$).
Similarly, Split CP using $\Bu$ is among the top performers on the 1354\_pegase dataset, but among the worst performers on the 118\_ieee dataset.
Such variability in performance further reinforces the value of model selection, which is most evident on the 1354\_pegase dataset where methods such as SCP with $\Bl$ and SFP CP perform well when the coverage is between 85\% and 90\%, but are out-performed by SCP with $\Bu$ when coverage is above 90\%.
In contrast, CPUL and CPUL-OMLT perform well across the entire coverage range.
This is explained by the fact that CPUL and CPUL-OMLT select the best construction \emph{for each dataset and target coverage $\alpha$}.

Finally, while CPUL and CPUL-OMLT perform similarly on the 89\_pegase and 118\_ieee datasets, CPUL-OMLT offers a clear improvement on the 1354\_pegase dataset, especially for coverage levels above 90\% (see Figure \ref{fig:cp_comparison}).
A more granular analysis of each model's behavior reveals that CPUL's performace is worse on samples where $[\Bl, \Bu]$ is small. Namely, CPUL achieves a coverage of only about 50\% across the 5\% of test samples with smallest initial interval $[\Bl, \Bu]$. Recall that these samples correspond to regions where the given bounds $\Bl, \Bu$ are the most accurate, which demonstrates the issue of paradoxical miscoverage (see Section \ref{sec:OMLT} and Figure \ref{fig:Paradoxical_Miscoverage}). CPUL-OMLT effectively mitigates this issue, which results in better coverage and smaller interval length overall.
\vspace{-7pt}
\section{Conclusion}
\vspace{-5pt}
This paper introduced CPUL-OMLT, a novel CP mechanism designed for settings where valid lower and upper bounds on the target variable are available. By integrating multiple interval construction strategies within the NCP framework, CPUL effectively leverages the structure of these bounds to improve efficiency. The OMLT mechanism leverages the strong condition that tight initial bounds already provide highly precise confidence intervals. Paradoxically, failure to account for this leads to undercoverage in the regions where the models perform best, a challenge that existing methods struggle to address.
Experimental validations were conducted on optimization problems across three power systems datasets, demonstrating that the proposed approach consistently outperforms the traditional CP methods by providing better efficiency.
The proposed CPUL-OMLT method provides state-of-the-art efficiency and remains consistent across different datasets, highlighting its robustness and practical relevance.

Future work could extend CPUL-OMLT beyond the efficient marginal coverage studied here. 
Potential directions include enforcing stronger conditional coverage to incorporate the already computed bounds $\Bl, \Bu$, integrating with optimization algorithms like branch and bound to directly boost downstream task performance, and adapting the method for high-dimensional/structured data to increase applicability in complex settings like energy systems.


\bibliography{iclr2026_conference}

@article{klamkin2025self,
  title={Self-Certifying Primal-Dual Optimization Proxies for Large-Scale Batch Economic Dispatch},
  author={Klamkin, Michael and Tanneau, Mathieu and Van Hentenryck, Pascal},
  journal={arXiv preprint arXiv:2510.15850},
  year={2025}
}

@article{grontas2025pinet,
  title={Pinet: Optimizing hard-constrained neural networks with orthogonal projection layers},
  author={Grontas, Panagiotis D and Terpin, Antonio and Balta, Efe C and D'Andrea, Raffaello and Lygeros, John},
  journal={arXiv preprint arXiv:2508.10480},
  year={2025}
}

@article{tordesillas2023rayen,
  title={Rayen: Imposition of hard convex constraints on neural networks},
  author={Tordesillas, Jesus and How, Jonathan P and Hutter, Marco},
  journal={arXiv preprint arXiv:2307.08336},
  year={2023}
}

@misc{miltenberger2025mipgap,
  author       = {Matthias Miltenberger},
  title        = {What is the MIPGap?},
  howpublished = {Gurobi Help Center},
  year         = {2025},
  month        = aug,
  note         = {Last updated August 28, 2025},
  url          = {https://support.gurobi.com/hc/en-us/articles/8265539575953-What-is-the-MIPGap},
  urldate      = {2026-01-15}
}

@article{wang2009comparison,
  title={A comparison of performance of several artificial intelligence methods for forecasting monthly discharge time series},
  author={Wang, Wen-Chuan and Chau, Kwok-Wing and Cheng, Chun-Tian and Qiu, Lin},
  journal={Journal of hydrology},
  volume={374},
  number={3-4},
  pages={294--306},
  year={2009},
  publisher={Elsevier}
}

@article{vivas2020systematic,
  title={A systematic review of statistical and machine learning methods for electrical power forecasting with reported mape score},
  author={Vivas, Eliana and Allende-Cid, H{\'e}ctor and Salas, Rodrigo},
  journal={Entropy},
  volume={22},
  number={12},
  pages={1412},
  year={2020},
  publisher={MDPI}
}

@article{broder2004network,
  title={Network applications of bloom filters: A survey},
  author={Broder, Andrei and Mitzenmacher, Michael},
  journal={Internet mathematics},
  volume={1},
  number={4},
  pages={485--509},
  year={2004},
  publisher={Taylor \& Francis}
}

@incollection{geoffrion2009lagrangean,
  title={Lagrangean relaxation for integer programming},
  author={Geoffrion, Arthur M},
  booktitle={Approaches to integer programming},
  pages={82--114},
  year={2009},
  publisher={Springer}
}

@article{pglib,  title={The {Power Grid Library} for Benchmarking {AC} Optimal Power Flow Algorithms},  author={Babaeinejadsarookolaee, Sogol and Birchfield, Adam and Christie, Richard D and Coffrin, Carleton and DeMarco, Christopher and Diao, Ruisheng and Ferris, Michael and Fliscounakis, Stephane and Greene, Scott and Huang, Renke and others},  journal={arXiv preprint arXiv:1908.02788},  year={2019}}

@article{pegase,  title={Contingency ranking with respect to overloads in very large power systems taking into account uncertainty, preventive, and corrective actions},  author={Fliscounakis, St{\'e}phane and Panciatici, Patrick and Capitanescu, Florin and Wehenkel, Louis},  journal={IEEE Transactions on Power Systems},  volume={28},  number={4},  pages={4909--4917},  year={2013},  publisher={IEEE}}

@misc{ieee_pstca,    title={Power systems test case archive},    author={{University of Washington, Dept. of Electrical Engineering}},    year={1999},    url={http://www.ee.washington.edu/research/pstca/},}

@misc{opf_generator,
  title = {{PGLearn.jl}: A Benchmark Suite for Optimal Power Flow Problems},
  author = {Tanneau, Mathieu and Klamkin, Michael},
  year = {2024},
  howpublished = {\url{https://github.com/AI4OPT/PGLearn.jl}}
}

@article{ml4opf,
  title={{PGL}earn--An Open-Source Learning Toolkit for Optimal Power Flow},
  author={Klamkin, Michael and Tanneau, Mathieu and Van Hentenryck, Pascal},
  journal={arXiv preprint arXiv:2505.22825},
  year={2025}
}

@inproceedings{cormode2018privacy,
  title={Privacy at scale: Local differential privacy in practice},
  author={Cormode, Graham and Jha, Somesh and Kulkarni, Tejas and Li, Ninghui and Srivastava, Divesh and Wang, Tianhao},
  booktitle={Proceedings of the 2018 International Conference on Management of Data},
  pages={1655--1658},
  year={2018}
}

@article{zhang2014these,
  title={These are not the k-mers you are looking for: efficient online k-mer counting using a probabilistic data structure},
  author={Zhang, Qingpeng and Pell, Jason and Canino-Koning, Rosangela and Howe, Adina Chuang and Brown, C Titus},
  journal={PloS one},
  volume={9},
  number={7},
  pages={e101271},
  year={2014},
  publisher={Public Library of Science San Francisco, USA}
}

@inproceedings{goyal2012sketch,
  title={Sketch algorithms for estimating point queries in nlp},
  author={Goyal, Amit and Daum{\'e} III, Hal and Cormode, Graham},
  booktitle={Proceedings of the 2012 joint conference on empirical methods in natural language processing and computational natural language learning},
  pages={1093--1103},
  year={2012}
}

@article{cormode2005improved,
  title={An improved data stream summary: the count-min sketch and its applications},
  author={Cormode, Graham and Muthukrishnan, Shan},
  journal={Journal of Algorithms},
  volume={55},
  number={1},
  pages={58--75},
  year={2005},
  publisher={Elsevier}
}

@inproceedings{charikar2002finding,
  title={Finding frequent items in data streams},
  author={Charikar, Moses and Chen, Kevin and Farach-Colton, Martin},
  booktitle={International Colloquium on Automata, Languages, and Programming},
  pages={693--703},
  year={2002},
  organization={Springer}
}

@article{oliveira2024split,
  title={Split conformal prediction and non-exchangeable data},
  author={Oliveira, Roberto I and Orenstein, Paulo and Ramos, Thiago and Romano, Joao Vitor},
  journal={Journal of Machine Learning Research},
  volume={25},
  number={225},
  pages={1--38},
  year={2024}
}

@article{chen2024real,
  title={Real-time risk analysis with optimization proxies},
  author={Chen, Wenbo and Tanneau, Mathieu and Van Hentenryck, Pascal},
  journal={Electric Power Systems Research},
  volume={235},
  pages={110822},
  year={2024},
  publisher={Elsevier}
}

@article{qiu2024dual,
  title={Dual conic proxies for AC optimal power flow},
  author={Qiu, Guancheng and Tanneau, Mathieu and Van Hentenryck, Pascal},
  journal={Electric Power Systems Research},
  volume={236},
  pages={110661},
  year={2024},
  publisher={Elsevier}
}

@article{klamkin2024dual,
  title={Dual Interior-Point Optimization Learning},
  author={Klamkin, Michael and Tanneau, Mathieu and Van Hentenryck, Pascal},
  journal={arXiv preprint arXiv:2402.02596},
  year={2024}
}

@article{chen2023end,
  title={End-to-end feasible optimization proxies for large-scale economic dispatch},
  author={Chen, Wenbo and Tanneau, Mathieu and Van Hentenryck, Pascal},
  journal={IEEE Transactions on Power Systems},
  year={2023},
  publisher={IEEE}
}

@article{sesia2023conformal,
  title={Conformal frequency estimation using discrete sketched data with coverage for distinct queries},
  author={Sesia, Matteo and Favaro, Stefano and Dobriban, Edgar},
  journal={Journal of Machine Learning Research},
  volume={24},
  number={348},
  pages={1--80},
  year={2023}
}

@article{lei2018distribution,
  title={Distribution-free predictive inference for regression},
  author={Lei, Jing and G’Sell, Max and Rinaldo, Alessandro and Tibshirani, Ryan J and Wasserman, Larry},
  journal={Journal of the American Statistical Association},
  volume={113},
  number={523},
  pages={1094--1111},
  year={2018},
  publisher={Taylor \& Francis}
}

@article{gupta2022nested,
  title={Nested conformal prediction and quantile out-of-bag ensemble methods},
  author={Gupta, Chirag and Kuchibhotla, Arun K and Ramdas, Aaditya},
  journal={Pattern Recognition},
  volume={127},
  pages={108496},
  year={2022},
  publisher={Elsevier}
}

@article{sesia2020comparison,
  title={A comparison of some conformal quantile regression methods},
  author={Sesia, Matteo and Cand{\`e}s, Emmanuel J},
  journal={Stat},
  volume={9},
  number={1},
  pages={e261},
  year={2020},
  publisher={Wiley Online Library}
}

@article{yang2024selection,
  title={Selection and Aggregation of Conformal Prediction Sets},
  author={Yang, Yachong and Kuchibhotla, Arun Kumar},
  journal={Journal of the American Statistical Association},
  pages={1--13},
  year={2024},
  publisher={Taylor \& Francis}
}

@article{liang2024conformal,
  title={Conformal prediction after efficiency-oriented model selection},
  author={Liang, Ruiting and Zhu, Wanrong and Barber, Rina Foygel},
  journal={arXiv preprint arXiv:2408.07066},
  year={2024}
}

@article{angelopoulos2021gentle,
  title={A gentle introduction to conformal prediction and distribution-free uncertainty quantification},
  author={Angelopoulos, Anastasios N and Bates, Stephen},
  journal={arXiv preprint arXiv:2107.07511},
  year={2021}
}

@book{vovk2005algorithmic,
  title={Algorithmic learning in a random world},
  author={Vovk, Vladimir and Gammerman, Alexander and Shafer, Glenn},
  volume={29},
  year={2005},
  publisher={Springer}
}

@article{romano2019conformalized,
  title={Conformalized quantile regression},
  author={Romano, Yaniv and Patterson, Evan and Candes, Emmanuel},
  journal={Advances in neural information processing systems},
  volume={32},
  year={2019}
}

@inproceedings{papadopoulos2002inductive,
  title={Inductive confidence machines for regression},
  author={Papadopoulos, Harris and Proedrou, Kostas and Vovk, Volodya and Gammerman, Alex},
  booktitle={Machine learning: ECML 2002: 13th European conference on machine learning Helsinki, Finland, August 19--23, 2002 proceedings 13},
  pages={345--356},
  year={2002},
  organization={Springer}
}

@article{carpentier1962contribution,
  title={Contribution to the economic dispatch problem},
  author={Carpentier, J},
  journal={Bulletin de la Societe Francoise des Electriciens},
  volume={3},
  number={8},
  pages={431--447},
  year={1962}
}

@inproceedings{tanneau2024dual,
title={Dual Lagrangian Learning for Conic Optimization},
author={Mathieu Tanneau and Pascal Van Hentenryck},
booktitle={The Thirty-eighth Annual Conference on Neural Information Processing Systems},
year={2024},
url={https://openreview.net/forum?id=gN1iKwxlL5}
}

@article{highs,
    title = {Parallelizing the dual revised simplex method},
    author = {Q. Huangfu and J. A. J. Hall},
    journal = {Mathematical Programming Computation},
    year = {2018},
    volume = {10},
    number = {1},
    pages = {119--142},
    doi = {10.1007/s12532-017-0130-5},
  }

@inproceedings{pytorch,
  author    = {Paszke, Adam and others},
  booktitle = {Advances in Neural Information Processing Systems 32},
  editor    = {Wallach, H. and Larochelle, H. and Beygelzimer, A. and d'Alché-Buc, F. and Fox, E. and Garnett, R.},
  pages     = {8024--8035},
  publisher = {Curran Associates, Inc.},
  title     = {{PyTorch: An Imperative Style, High-Performance Deep Learning Library}},
  year      = {2019}
}

@misc{lightning,
author = {Falcon, William and {The PyTorch Lightning team}},
doi = {10.5281/zenodo.3828935},
license = {Apache-2.0},
month = mar,
title = {{PyTorch Lightning}},
url = {https://github.com/Lightning-AI/lightning},
version = {1.4},
year = {2019}
}

@book{boyd2004convex,
  title={Convex optimization},
  author={Boyd, Stephen P and Vandenberghe, Lieven},
  year={2004},
  publisher={Cambridge university press}
}

@book{wolsey1999integer,
  title={Integer and combinatorial optimization},
  author={Wolsey, Laurence A and Nemhauser, George L},
  year={1999},
  publisher={John Wiley \& Sons}
}

@article{scavuzzo2024learning,
  title={Learning optimal objective values for MILP},
  author={Scavuzzo, Lara and Aardal, Karen and Yorke-Smith, Neil},
  journal={arXiv preprint arXiv:2411.18321},
  year={2024}
}

@article{zhang2021convex,
  title={A convex neural network solver for DCOPF with generalization guarantees},
  author={Zhang, Ling and Chen, Yize and Zhang, Baosen},
  journal={IEEE Transactions on Control of Network Systems},
  volume={9},
  number={2},
  pages={719--730},
  year={2021},
  publisher={IEEE}
}

@article{rosemberg2024learning,
  title={Learning Optimal Power Flow value functions with input-convex neural networks},
  author={Rosemberg, Andrew and Tanneau, Mathieu and Fanzeres, Bruno and Garcia, Joaquim and Van Hentenryck, Pascal},
  journal={Electric power systems research},
  volume={235},
  pages={110643},
  year={2024},
  publisher={Elsevier}
}

@article{mak1999monte,
  title={Monte Carlo bounding techniques for determining solution quality in stochastic programs},
  author={Mak, Wai-Kei and Morton, David P and Wood, R Kevin},
  journal={Operations research letters},
  volume={24},
  number={1-2},
  pages={47--56},
  year={1999},
  publisher={Elsevier}
}
\bibliographystyle{iclr2026_conference}
\newpage
\appendix
\section{Experiment Details}
\label{sec:Experiment Details}
\subsection{Problem formulations}
\label{sec:background:optimization}

With a slight abuse of notation for the readability, here equality constraints are written out explicitly.
Let $\mathbf{S}(x)$ denote the feasible set of, i.e, 
\begin{align*}
    \mathbf{S}(x) = \{s \in \mathbb{R}^{n} | g_{x}(s) \leq 0, h_{x}(s) = 0 \}
\end{align*}
A candidate solution $s \in \mathbb{R}^{n}$  is \emph{primal feasible} if $s \in \mathbf{S}(x)$, i.e., if $g_{x}(s) \leq 0$ and $h_{x}(s) = 0$, otherwise it is \emph{infeasible}. 
Suppose that there exists a unique optimal solution $x^{*}$. Then, $f_x(x^{*}) \leq f_x(s)$ for all $s \in \mathbf{S}(x)$.\\

The Lagrangian is defined as:\begin{align}
    \mathcal{L}({s}, \lambda, {\mu}) = f_x({s}) + \lambda^T g_x({s}) +  \mu^T h_x({s}),
\end{align}where $\lambda \in \mathbb{R}^m$ and $\mu \in \mathbb{R}^p$ are Lagrange multipliers. Its corresponding dual problem is defined as 
\begin{subequations}
\begin{align}
    \label{eq:dual_optimization}
    \varphi(x) = \max_{\lambda,\mu} \quad &\inf_{{s}} \mathcal{L}({s}, {\lambda}, {\mu}) \\
    \text{s.t.} \quad &\lambda \geq 0
\end{align}
\end{subequations}
Dual feasibility is defined similarly to primal feasibility, $\Lambda(\lambda)=\{\lambda\in\mathbb{R}^m \vert\lambda\geq0\}$. Suppose there exists a unique optimal dual solution $(\lambda_x, \mu_x)$. By the weak duality theorem, we must have $\Phi(x) \geq \varphi(x)$.

\paragraph{Economic Dispatch}
The experiments evaluate the proposed methods in the context of the Optimal Power Flow (OPF) problem \cite{carpentier1962contribution}, a fundamental challenge in power system operations that focuses on optimizing generation dispatch while satisfying various physical and engineering constraints. Specifically, we address the Economic Dispatch problem with soft thermal constraints. In this section, we present the exact mathematical formulations used for learning the primal and dual proxies, following \cite{chen2023end,klamkin2024dual}.

Primal formulation (equivalent to ``EconomicDispatch'' with ``soft\_thermal\_limit'' enabled in \citet{opf_generator}):
\begin{subequations} \label{eq:ed primal_optimization}
    \begin{align}
        \min_{p,f,\xi} \quad & c^{\top} p + M e^{\top} \xi \label{eq:primal objective}\\
        s.t. \quad
        & e^{\top}p = e^{\top} d && [\lambda]\\
        & \Phi A_{g} p - f = \Phi A_{d} d && [\pi]\\
        & {\phantom{+}} f + \xi \geq \underline{f} && [\underline{\mu}]\\
        & {-} f + \xi \geq -\bar{f} && [\bar{\mu}]\\
        & \underline{p} \leq p \leq \bar{p} && [\underline{z}, \bar{z}]\\
        & \xi \geq 0 && [y] 
    \end{align}
\end{subequations}
    where $p$ is the vector of generation, $d$ is the vector of demand, and $f$ is the vector of power flows.
    The vector $\xi \in \mathbb{R}^{E}$ denotes the vector of thermal violations.
    Matrix $\Phi \in \mathbb{R}^{E \times N}$ is the nodal PTDF matrix, $A_{g} \in \mathbb{R}^{N \times G}$ is the incidence matrix of generators, and $A_{d}\in \mathbb{R}^{N \times D}$ is the incidence matrix of loads.

    The dual problem reads
    \begin{subequations}\label{eq:dual_optimization-a}
    \begin{align}
        \max_{\lambda, \pi, \mu} \quad
            & \lambda e^{\top}d 
            + (\Phi A_{d} d)^{\top} \pi
            + \underline{f}^{\top} \underline{\mu} - \bar{f}^{\top}\bar{\mu} 
            + \underline{p}^{\top} \underline{z} - \bar{p}^{\top} \bar{z}\label{eq:dual objective}\\
        s.t. \quad 
            & \lambda e + (\Phi A_{g})^{\top} \pi + \underline{z} - \bar{z} = c\\
            & -\pi + \underline{\mu} - \bar{\mu} = 0\\
            & \underline{\mu} + \bar{\mu} + y = Me\\
            & \underline{\mu}, \bar{\mu}, \underline{z}, \bar{z}, y \geq 0
    \end{align}
    \end{subequations}

\subsection{Data generation}
\label{data123}
    Note that all CP methods require true labels to compute exact residuals during the calibration step \eqref{eq:NCP:conformalization}, and the test set must be labeled as well for performance evaluation. 
    These true labels are computed using the LP solver \mbox{HiGHS~\cite{highs}}, via PGLearn \cite{opf_generator}. 
    For the significantly larger training set $\mathcal{D}_{\mathrm{train}}$, true labels are not required, as the training process is self-supervised. 
    The three datasets used in this study are based on selected snapshots from the Power Grid Lib - Optimal Power Flow collection \cite{pglib}.
    The sampling distribution is consistent across all cases. For each sample \(i\), each load \(d_l\) is generated as 
    \[
    d_l^{(i)} = \alpha^{(i)}\beta_l^{(i)}d_l^0,
    \]
    where \(\alpha^{(i)}\) represents a ``global'' factor and \(\beta_l^{(i)}\) is a ``local'' factor. 
    The global factor \(\alpha^{(i)}\) follows a \(\text{Uniform}(0.6, 1.0)\) distribution for 89\_pegase and 118\_ieee, and a \(\text{Uniform}(0.8, 1.05)\) distribution for 1354\_pegase. 
    The local factor \(\beta_l^{(i)}\) is sampled from \(\text{Uniform}(0.85, 1.15)\) in all cases.
    Each snapshot is described next.

\subsubsection{89\_pegase}
\label{data3}
This case accurately represents the size and complexity of a portion of the European high-voltage transmission network. The network comprises 89 buses, 12 generators, and 210 branches, operating at 380, 220, and 150 kV. The data originate from the Pan European Grid Advanced Simulation and State Estimation (PEGASE) project, which was part of the 7th Framework Program of the European Union~\cite{pegase}.

\subsubsection{118\_ieee}\label{data2}
The test case represents a standard benchmark in power systems engineering, modeling a large-scale electric grid inspired by the American Electric Power system in the Midwestern United States as of December 1962. It includes 118 buses, multiple generators, loads, and transmission lines, and is extensively used by researchers to analyze power system operations under various conditions~\cite{ieee_pstca}.

\subsubsection{1354\_pegase}\label{data1}
The data originate from the Pan European Grid Advanced Simulation and State Estimation (PEGASE) project, which is part of the 7th Framework Program of the European Union. This case accurately represents the size and complexity of a segment of the European high voltage transmission network. The network comprises 1,354 buses, 260 generators, and 1,991 branches, operating at 380 and 220 kV~\cite{pegase}.

\subsection{Optimization proxies (base models)}
\label{sec:background:proxies}
Optimization proxies are efficient and scalable neural network (NN) models that approximate the input-output mapping of optimization solvers. 
For instance, when using NN models, $\theta$ denotes the weights of the NN.
Predicting an optimal solution for the primal problem consists of training a model $\hat{\mathcal{M}}^p_{\theta}(x)= \hat{s} \in \mathbb{R}^{n}$ such that $\hat{s}$ is approximating the true optimal solution of the problem parameterized by $x$. Denote the corresponding estimated primal objective value as $\hat{\Phi}(x) = f(\hat{\mathcal{M}}^p_{\theta}(x))$; similarly, the estimated dual objective value is denoted $\hat{\varphi}(x) = \inf_{{s}} \mathcal{L}({s}, \hat{\mathcal{M}}^d_{\theta}(x))$.

In this paper, all the primal proxies $\hat{\mathcal{M}}^p_{\theta}(x)$ are assumed to be primal-feasible, and all the dual proxies are assumed to be dual-feasible. The predicted objective values \(\{\hat{\Phi}(X_i)\}_{i \in I_{cal} \cup I_{test}}\) and \(\{\hat{\varphi}(X_i)\}_{i \in I_{cal} \cup I_{test}}\) are recovered from the proxies by evaluating \eqref{eq:primal objective} and \eqref{eq:dual objective}, respectively. Note that, by the duality theorem, for all \(i \in I_{cal}\), the following must hold:  
\[
    \hat{\Phi}(X_i) \leq \Phi(X_i) \leq \hat{\varphi}(X_i).
\]

To ensure feasibility, the following strategies are employed, as detailed in the subsequent subsections.

\subsubsection{Primal feasible proxies}
    A primal-feasible solution to \ref{eq:ed primal_optimization} can be obtained by following the procedure below, similar to \cite{chen2023end}:
    \begin{enumerate}
        \item Predict $\tilde{p}\in[\underline{p},\overline{p}]^G$.
        \item Use the power balance layer \citep[Eq.~4]{chen2023end} to obtain $p$ from $\tilde{p}$ such that $e^\top p = e^\top d$ and $\underline{p}\leq p\leq \overline{p}$.
        \item Recover $f=\Phi A_g p - \Phi A_d d$
        
        \item Recover $\xi=\max\big(\max(0, f-\overline{f}),\, \max(0,\underline{f}-f)\big)$
    \end{enumerate}
\subsubsection{Dual feasible proxies}
    The Dual Lagrangian Learning framework \cite{tanneau2024dual} is applied; 
    the specific dual recovery procedure reads as follows:
    \begin{enumerate}
        \item Predict $\lambda \in \mathbb{R}$ and $\pi \in [-M, M]^{E}$.
        \item Recover $\underline{\mu} = \max(0, \pi)$ and $\bar{\mu} = \max(0, -\pi)$
        \item Set $z = c - \lambda e - (\Phi A_{g})^{\top} \pi$
        \item Recover $\underline{z} = \max(0, z)$ and $\bar{z} = \max(0, -z)$
    \end{enumerate}
\subsubsection{Details of training}

Both proxies are trained using the self-supervised approach outlined in \cite{chen2023end,tanneau2024dual,klamkin2024dual}. The network architecture is a feed-forward network with softplus activations and a 5\% dropout rate. 
The model training is implemented using the ML4OPF \cite{ml4opf} library which itself is based on PyTorch \cite{pytorch} and Lightning \cite{lightning}.
Comprehensive hyperparameter tuning is performed, optimizing learning rate, decay strategy, and model architecture, with the best configuration selected. 
The optimal hyper parameters are presented in Table \ref{tab:proxy_configurations}.

For all datasets, both the primal and dual proxies are implemented using the \textit{softplus} activation function. The configurations and performance metrics are summarized in Table~\ref{tab:proxy_configurations}.
\begin{table*}[!t]
    \centering
    \small 
    \caption{Proxy Configurations and Performance Metrics}
    \label{tab:proxy_configurations}
    \begin{tabular*}{\textwidth}{@{\extracolsep{\fill}}lcccccc@{}}
        \toprule
        \textbf{Dataset} & \textbf{Proxy} & \textbf{\#Layers} & \textbf{\#Units/Layer} & \textbf{Learning Rate} & \textbf{Decay (Rate/Steps)} & \textbf{MAPE (\%)} \\
        \midrule
        89\_pegase & Primal & 3 & 128 & 0.001 & 0.9 / 15 & 0.23 \\
                   & Dual   & 4 & 128 & 0.05  & 0.7 / 15 & 0.18 \\
        \midrule
        118\_ieee  & Primal & 3 & 256 & 0.05  & 0.9 / 20 & 0.19 \\
                   & Dual   & 4 & 256 & 0.05  & 0.7 / 15 & 0.10 \\
        \midrule
        1354\_pegase & Primal & 4 & 2048 & 0.001  & 0.75 / 15 & 3.06 \\
                     & Dual   & 4 & 2048 & 0.0001 & 0.85 / 10 & 0.97 \\
        \bottomrule
    \end{tabular*}
\end{table*}

Note that the hyperparameters for 1354\_pegase result in a relatively high primal MAPE, primarily due to the sensitivity of optimal hyperparameters in one of the ten splits, which reflects a commonly observed challenge in robust hyperparameter tuning within deep learning.


\subsection{Experiment Details}\label{sec:Proxies used in experiments} All experiments are conducted on RHEL9 machines with 24 Intel Xeon 2.7 GHz CPU cores,  equipped with an NVIDIA V100 GPU.
Random shuffling and splitting of data samples is performed separately for each dataset. The process is repeated 10 times for statistical reliability of the results, and the average and standard deviation of the results are reported. 
For each round, 40,000 samples are used for training, 5,000 for calibration, and 5,000 for testing, ensuring robust evaluation.
\subsubsection{Details of comparison methods}
\label{app:experiments:cp_baselines}
CPUL and its optimized version, CPUL-OMLT, are compared against three other CP methods. For clarity, the construction of all methods used in the experiments is summarized in Table~\ref{tab:proxy_configurations}. 

The first comparison method is the widely applied Split CP \cite{vovk2005algorithmic}, as reviewed in Section~\ref{sec:Background}. With two base models, $\hat{B}^u$ and $\hat{B}^l$, classical CP is independently applied to each model. Specifically, the unsigned residuals \(\hat{s}^u(x, y) = y - \hat{B}^u(x)\) and \(\hat{s}^l(x, y) = y - \hat{B}^l(x)\) serve as score functions to capture residual distribution differences between $\hat{B}^u$ and $\hat{B}^l$. When a single base model constructs prediction intervals, the other is incorporated via a post-processing step, as described in \eqref{eq:strengthening}, aligning the Split CP constructions with CPUL-u and CPUL-l.

The second category of comparison focuses on CQR methods \cite{romano2019conformalized, sesia2020comparison}. In the adopted versions, the fitted quantile estimates are replaced with the base models: the upper quantile is substituted with $\hat{B}^l$ and the lower quantile with $\hat{B}^u$. For CQR-r, an additional scaling factor of \(1/(\hat{B}^u - \hat{B}^l)\) is applied. 



\subsubsection{OMLT implementation details}
\label{sec:CPUL and OMLT implementation details}

Note that OMLT requires optimizing the parameter $\ell$. 
In the paper's implementation, this is achieved by performing a grid search on $\ell$.
This grid search is performed using 1000 samples, reserved from the calibration set. For each CPUL submethod, choose a hyper-parameter $\ell$ using the following strategy. For each \(\ell\) in the search space, \eqref{eqn:OMLT} is solved by applying \eqref{eq:NCP:conformalization} to the reserved 1000 samples at the confidence level of \(1 - \alpha\) coverage. 
The optimal \(\ell\) is selected based on the average prediction interval length at the given \(1-\alpha\) level on the hold-out set of 1,000 samples. Then, each CPUL submethod with the optimal \(\ell\), calibration (and model-selection) is conducted using the remaining 4,000 samples. Lastly, CPUL-OMLT is verified on the test set.

\section{Additional experiment 
results}\label{sec:Additional experiment results}

See additional details of experiments given in Table \ref{table full:results}.
\begin{table*}[!h]
    \centering
    \caption{Comparisons of CP Methods Across Different Datasets: 89\_pegase, 118\_ieee, and 1354\_pegase.}
    \label{table full:results}
    \resizebox{\textwidth}{!}{
    \begin{tabular}{c|rr|rr|rr|rr|rr|rr}
        \toprule
        
                \cmidrule(lr){2-13}
        \textbf{UQ Methods} 
            & \multicolumn{2}{c}{$\alpha = 2.5\%$} 
            & \multicolumn{2}{c}{$\alpha = 5\%$} 
            & \multicolumn{2}{c}{$\alpha = 7.5\%$}
            & \multicolumn{2}{c}{$\alpha = 10\%$}
            & \multicolumn{2}{c}{$\alpha = 12.5\%$}
            & \multicolumn{2}{c}{$\alpha = 15\%$} \\
        \cmidrule(lr){2-3} \cmidrule(lr){4-5} \cmidrule(lr){6-7} \cmidrule(lr){8-9} \cmidrule(lr){10-11} \cmidrule(lr){12-13}
            & \textbf{PICP (\%)} & \textbf{Length (\%)} 
            & \textbf{PICP (\%)} & \textbf{Length (\%)} 
            & \textbf{PICP (\%)} & \textbf{Length (\%)} 
            & \textbf{PICP (\%)} & \textbf{Length (\%)}
            & \textbf{PICP (\%)} & \textbf{Length (\%)}
            & \textbf{PICP (\%)} & \textbf{Length (\%)} \\
        \midrule
        & \multicolumn{12}{c}
        {\textbf{89\_pegase}} \\ 
                \midrule
        $[\hat{B}^l, \hat{B^u}]$ & 100.0 (0.00) & 0.410 (0.016)&100.0 (0.00) & 0.410 (0.016)&100.0 (0.00) & 0.410 (0.016)&100.0 (0.00) & 0.410 (0.016)&100.0 (0.00) & 0.410 (0.016)&100.0 (0.00) & 0.410 (0.016)\\
        \cmidrule(lr){2-13}
            Split CP w/ $\hat{B}^l$ & 	97.56(0.28)
 & \color{red}{0.281(0.027)}
&95.02(0.40) & \color{red}{0.242(0.026)}
 
 & 	92.61(0.57)
 & 	\color{red}{0.218(0.015)}
 &90.15(0.60)
 &	0.197(0.007)
 & 87.91(0.85)
 & 0.188(0.006) &85.50(1.03)&0.181(0.007)
\\
 Split CP w/ $\hat{B}^u$ & 97.57(0.26)
 & 0.260(0.019)
 & 95.02(0.43)
 & 	0.230(0.006)
 &	92.40(0.69)
 & {0.215(0.006)}
 & 	90.02(0.54)
& 	\color{red}{0.205(0.007)}
&87.68(0.52)
&	\color{red}{0.194(0.008)}
&85.09(0.44) &	\color{red}{0.184(0.008)}
  \\
SFD CP& 98.18(0.05) & \color{blue}{0.239(0.006)} & 95.97(0.27) & \color{blue}{0.215(0.009)} & 93.61(0.35) & \color{blue}{0.198(0.007)} & 91.23(0.56) & \color{blue}{0.188(0.007)} &88.85(0.69)
&\color{blue}{0.178(0.006)} & 86.63(0.84)&	\color{blue}{0.167(0.006)}
 \\
CQR & 		98.14(0.50)
 & \color{red}{0.402(0.018)}
 & 95.95(1.05)
 &\color{red}{0.398(0.018)}
 & 93.78(1.49)
 & 	\color{red}{0.394(0.018)}
 & 	91.40(1.80)
 & \color{red}{0.389(0.018)} &89.54(2.24)&\color{red}{0.384(0.017)}&87.23(2.50) &\color{red}{0.381(0.017)}
 \\
        
 CQR-r& 	97.79(0.40)
& 	\color{red}{0.376(0.019)}
 & 	95.37(0.46)

 & \color{red}{0.359(0.020)}
 & 	93.09(0.55)
& \color{red}{0.344(0.020)}
 & 	90.71(0.72)
 & \color{red}{0.332(0.020)}
 &88.24(0.72) &\color{red}{0.319(0.020)} &85.48(0.86)&\color{red}{0.309(0.019)}\\
 
        
        
CPUL (ours) & 	98.18(0.05)
 & 	\color{blue}{0.239(0.006)}
 & 95.97(0.27)
 & \color{blue}{0.215(0.009)}
 & 93.61(0.35)
 &	\color{blue}{0.198(0.007)}
 & 	91.23(0.56)
 & \color{blue}{0.188(0.007)} &88.85(0.69) &\color{blue}{0.178(0.006)} &86.63(0.84)&\color{blue}{0.167(0.006)}
 \\
CPUL-OMLT (ours) & 98.18(0.05) & \color{blue}{0.239(0.006)} & 95.97(0.27) & \color{blue}{0.215(0.009)} & 93.61(0.35)
 & \color{blue}{0.198(0.007)} & 90.28(0.51) & \color{blue}{0.187(0.008)} &88.08(0.90)&\color{blue}{0.175(0.010)}&85.57(1.07) &\color{blue}{0.165(0.009)}\\
        \midrule
        & \multicolumn{12}{c}
        {\textbf{118\_ieee}} \\ 
        \midrule
        $[\hat{B}^l, \hat{B^u}]$ & 100.0 (0.00)& 0.281 (0.085)&100.0 (0.00)& 0.281 (0.085)&100.0 (0.00)& 0.281 (0.085)&100.0 (0.00)& 0.281 (0.085)&100.0 (0.00)& 0.281 (0.085)&100.0 (0.00)& 0.281 (0.085)\\
        \cmidrule(lr){2-13}
       Split CP w/ $\hat{B}^l$ 
       & 	97.68(0.29)
 & 	\color{red}{0.242(0.149)}
&95.30(0.48) & \color{red}{0.230(0.148)}
 
 & 	92.75(0.53)
 & 	\color{red}{0.217(0.147)}
 & 90.34(0.63)
 &	\color{red}{0.206(0.146)}
 & 87.78(0.58)
 & \color{red}{0.196(0.143)} &85.35(0.69) &\color{red}{0.186(0.139)}
\\
 Split CP w/ $\hat{B}^u$ & 97.75(0.15)
 & 	\color{blue}{0.144(0.007)}
 & 95.39(0.26)
 & 		\color{blue}{0.127(0.005)}
 &	92.74(0.50)
 & 	\color{blue}{0.115(0.004)}
 & 	90.00(0.50)
& 		\color{blue}{0.105(0.004)}
&87.43(0.55)
&	\color{blue}{0.098(0.005)}
&84.92(0.52) &	\color{blue}{0.091(0.005)}
  \\
SFD CP& 97.68(0.25) & 0.180(0.007) & 95.19(0.30) & 0.162(0.008) & 92.72(0.47)& 0.146(0.011) & 90.10(0.50) & 0.135(0.013) &87.56(0.69)
& 0.125(0.015) & 85.13(0.66) &0.118(0.017)
 \\
CQR & 		97.76(0.18)
 & \color{red}{0.214(0.146)}
 & 	95.57(0.69)
 &\color{red}{0.205(0.147)}
 & 	93.41(1.16)
 & 	\color{red}{0.198(0.147)}
 & 	91.34(1.66)
 & \color{red}{0.193(0.147)} &90.08(2.75)&\color{red}{0.190(0.147)} &87.17(2.56)&\color{red}{0.184(0.148)}
 \\
        
 CQR-r& 	{97.83(0.40)}
& 		\color{red}{0.206(0.166)}
 & 	{95.62(0.78)}

 & \color{red}{0.196(0.168)}
 & 	93.05(0.92)
&\color{red}{0.187(0.169)}
 & 	90.88(1.33)
 & 	\color{red}{0.180(0.167)}
 &89.56(2.49) &\color{red}{0.176(0.163)} &86.58(2.29) &\color{red}{0.166(0.157)}\\
 
        
        
CPUL (ours) & 	97.81(0.18)
 & 		\color{blue}{0.143(0.007)}
 & 95.45(0.30)
 & 	\color{blue}{0.127(0.005)}
 & 92.83(0.53)
 &		\color{blue}{0.115(0.005)}
 & 	90.02(0.46)
 & 	\color{blue}{0.105(0.004)} &87.43(0.53)&	\color{blue}{0.098(0.005)}&85.01(0.53) &	\color{blue}{0.091(0.005)}
 \\
CPUL-OMLT (ours) & 97.77(0.15) & 	\color{blue}{0.143(0.008)} &95.41(0.23) & 	\color{blue}{0.126(0.006)} & 92.78(0.56)
 & 	\color{blue}{0.114(0.006)} & 90.01(0.48) & 	\color{blue}{0.103(0.007)} &87.50(0.59)&	\color{blue}{0.095(0.006)}&84.93(0.62) &	\color{blue}{0.089(0.006)}\\ \midrule
        & \multicolumn{12}{c}
        {\textbf{1354\_pegase }} \\ 
         \midrule $[\hat{B}^l, \hat{B^u}]$ &
        100.0 (0.00) &3.949 (3.807)&
        100.0 (0.00) &3.949 (3.807)&
        100.0 (0.00) &3.949 (3.807)&
        100.0 (0.00) &3.949 (3.807)&
        100.0 (0.00) &3.949 (3.807)&
        100.0 (0.00) &3.949 (3.807)\\
        \midrule
        \cmidrule(lr){2-13}
            Split CP w/ $\hat{B}^l$ & 	97.31(0.35)
 & 	\color{blue}{1.560(1.267)}
&94.64(0.32) & \color{blue}{1.324(1.212)}
 & 	92.05(0.40)
 & 	\color{blue}{1.363(1.186)}
 & 89.60(0.54)
 &	\color{blue}{1.308(1.160)}
 & 87.07(0.76)
 & \color{red}{1.262(1.135)} &84.64(0.69) &\color{red}{1.225(1.123)}
\\
 Split CP w/ $\hat{B}^u$ & 97.14(0.31)
 & 	\color{red}{3.568(3.837)}
 & 94.66(0.37)
 & 	\color{red}{2.680(2.724)}
 &	92.15(0.48)
 & \color{red}{2.207(1.678)}
 & 	89.40(0.49)
& 	\color{red}{1.570(0.853)}
&87.03(0.57)
&{1.073(0.328)}
&84.75(0.55) &\color{blue}{0.686(0.092)}
  \\
SFD CP& 97.45(0.10) & 2.975(2.598) & 95.01(0.25) & 2.324(2.037) & 92.38(0.21) & 1.741(1.280) & 90.22(0.27) & 1.456(0.609) &87.74(0.35)
& \color{blue}{0.837(0.069)} & 85.09(0.51) &0.718(0.050)
 \\
CQR & 		97.24(0.32)
 & \color{red}{3.590(3.947)}
 & 	94.75(0.36)
 &\color{red}{3.552(3.943)}
 & 	92.45(0.44)
 & \color{red}{	3.524(3.942)}
 & 	90.13(0.43)
 & \color{red}{3.501(3.939)} &87.54(0.62) &\color{red}{3.480(3.938)} &84.87(0.75) &\color{red}{3.461(3.937)}
 \\
        
 CQR-r& 	{97.48(0.31)}
& 	\color{red}{3.570(3.870)}
 & 	{95.37(0.65)}
 & \color{red}{3.406(3.947)}
 & 	{93.51(0.83)}
& \color{red}{3.322(3.981)}
 & 	{91.57(1.32)}
 & 	\color{red}{3.271(3.997)}
 &{89.92(2.11)} &\color{red}{3.238(4.003)}&{88.25(3.09)} &\color{red}{3.212(4.006)}\\
 
        
        
CPUL (ours) & 	97.25(0.28)
 & 	\color{blue}{1.554(1.275)}
 & 94.62(0.33)
 & \color{blue}{1.321(1.215)}
 & 92.14(0.42)
 &	\color{blue}{1.360(1.189)}
 & 	89.65(0.56)
 & \color{blue}{1.306(1.162)} &87.74(0.35) &\color{blue}{0.837(0.069)} &84.80(0.57) &\color{blue}{0.632(0.048)}
 \\
CPUL-OMLT (ours) 
    & 97.31(0.22) & \color{blue}{1.205(0.794)}
    & 94.66(0.42) & \color{blue}{1.156(0.896)}
    & 92.26(0.54) & \color{blue}{1.108(0.926)}
    & 89.69(0.49) & \color{blue}{1.037(0.890)}
    & 87.74(0.35) & \color{blue}{0.837(0.069)} 
    & 85.68(2.50) & \color{blue}{0.642(0.114)}\\
        \bottomrule
\end{tabular}}
\begin{tablenotes}
    \scriptsize 
    \item * For each dataset and $\alpha$ value, the three shortest intervals are colored blue, while the three largest intervals are colored red.
\end{tablenotes}

\end{table*}

\end{document}